\documentclass{article}

\PassOptionsToPackage{numbers}{natbib}
\usepackage[preprint]{neurips_2026}

\usepackage[utf8]{inputenc} 
\usepackage[T1]{fontenc}    
\usepackage{hyperref}       
\usepackage{url}            
\usepackage{booktabs}       
\usepackage{amsfonts}       
\usepackage{nicefrac}       
\usepackage{microtype}      
\usepackage{xcolor}         

\usepackage[textsize=tiny]{todonotes}
\usepackage{amsmath}
\usepackage{amssymb}
\usepackage{titletoc}
\usepackage{mathtools}
\usepackage{multirow}
\usepackage{amsthm}
\usepackage{subcaption}
\usepackage{colortbl}   
\usepackage{minitoc}
\usepackage{dsfont}
\usepackage{float}
\usepackage[most]{tcolorbox}
\usepackage{enumitem} 
\definecolor{mytheme}{RGB}{0, 51, 102} 
\definecolor{TableBlue}{RGB}{235, 245, 245}
\usepackage{cleveref}
\usepackage{wrapfig}
\usepackage{listings}

\definecolor{addgreen}{RGB}{0,120,60}
\definecolor{delred}{RGB}{180,40,40}
\definecolor{codegray}{RGB}{245,245,245}

\newtcolorbox{rqbox}[1][]{
  enhanced, 
  attach boxed title to top left={xshift=10pt, yshift*=-\tcboxedtitleheight/2}, 
  colback=mytheme!5!white, 
  colframe=mytheme,        
  coltitle=white,          
  title={Research Questions}, 
  fonttitle=\bfseries\sffamily,
  boxed title style={
    colback=mytheme,      
    arc=3pt,             
    outer arc=3pt,
    boxrule=0pt,
  },
  arc=3pt,                
  boxrule=1pt,           
  left=10pt, right=10pt, top=15pt, bottom=10pt, 
  #1
}

\newcommand{\hltext}[1]{\textcolor{mytheme}{\textbf{#1}}}

\title{Flexible Entropy Control in RLVR from a Gradient-Preserving Perspective}

\author{%
	Kun Chen\textsuperscript{1,2}, \quad
	Peng Shi\textsuperscript{3,*},  \quad
	Fanfan Liu\textsuperscript{3},  \\
    \textbf{Haibo Qiu}\textsuperscript{3}, \quad
	\textbf{Zhixiong Zeng}\textsuperscript{3,2}, \quad
    \textbf{Siqi Yang\textsuperscript{3}},  \quad
    \textbf{Wenji Mao\textsuperscript{2,1,*}}\\
	\textsuperscript{1}School of Artificial Intelligence, University of Chinese Academy of Sciences \\
	\textsuperscript{2}MAIS, Institute of Automation, Chinese Academy of Sciences, \quad
	\textsuperscript{3}Meituan \\
	\texttt{\{chenkun2024,wenji.mao\}@ia.ac.cn}  \quad
	\texttt{shipeng10@meituan.com}
}

\begin{document}

\maketitle
\begingroup
\renewcommand\thefootnote{}\footnotetext{%
	\textsuperscript{*}\;Corresponding authors.
}
\addtocounter{footnote}{-1}
\endgroup

\begin{abstract}
Reinforcement Learning with Verifiable Rewards (RLVR) has emerged as a critical method for enhancing the reasoning capabilities of Large Language Models (LLMs). However, continuous training often leads to policy entropy collapse, characterized by a rapid decay in entropy that results in premature overconfidence, reduced output diversity, and vanishing gradient norms that inhibit learning. Gradient-Preserving Clipping is a primary factor influencing these dynamics, but existing mitigation strategies are largely static and lack a framework connecting clipping mechanisms to precise entropy control. This paper proposes reshaping entropy control in RL from the perspective of Gradient-Preserving Clipping. We first theoretically and empirically verify the contributions of specific importance sampling ratio regions to entropy growth and reduction. Leveraging these findings, we introduce a novel regulation mechanism using dynamic clipping thresholds to precisely manage entropy. Furthermore, we design and evaluate dynamic entropy control strategies, including increase-then-decrease, decrease-increase-decrease, and oscillatory decay. Experimental results demonstrate that these strategies effectively mitigate entropy collapse and achieve superior performance across multiple benchmarks.
\end{abstract}

\section{Introduction}
Reinforcement Learning with Verifiable Rewards (RLVR) \cite{lambert2024tulu,guo2025deepseek} has become an important training paradigm for boosting the reasoning capabilities of Large Language Models (LLMs) across diverse applications. As a representative algorithm of RLVR, the Group Relative Policy Optimization (GRPO) \cite{shao2024deepseekmath} has been widely adopted and proven effective in enhancing LLM reasoning \cite{guo2025deepseek, abdin2025phi, yang2025qwen3, zeng2025simplerlzooinvestigatingtamingzero, bercovich2025llama, zhang2025r1vllearningreasonmultimodal}. Nonetheless, a growing body of research also indicates that uncontrolled continuous training in RLVR may drive LLMs towards policy entropy collapse \cite{cui2025entropymechanismreinforcementlearning, shen2025entropycontrolllmrlalgorithms, cheng2025reasoningexplorationentropyperspective, yu2025dapoopensourcellmreinforcement}, manifested as a rapid decay in entropy to near-zero values in the training process.

Entropy collapse in LLMs poses several challenges to RLVR research. On one hand, a precipitous drop in policy entropy in early stages of RL training causes the model to become prematurely overconfident in its outputs, at the risk of sacrificing output diversity and thus being trapped in a locally optimal solution. On the other hand, \citet{shen2025entropycontrolllmrlalgorithms} has shown that the norm of the model's training gradients during the RL optimization is constrained by policy entropy, which inhibits the continuous upgrade of the model in later stages of training and  ultimately affects model performance. A significant factor influencing policy entropy dynamics during RLVR training is Gradient-Preserving Clipping. Originated from Clipped Proximal Policy Optimization (PPO-Clip) \cite{schulman2017proximal}, this method establishes upper and lower clipping bounds on the importance sampling ratio to optimize the policy within a trust region. 


Related studies on RLVR believe that, although harshly clipping tokens outside this trust region can stabilize the training process to a certain extent, it also impairs the diversity of model outputs due to ignoring some low-probability points, leading to a continuous decline in entropy and even entropy collapse \cite{yu2025dapoopensourcellmreinforcement, minimax2025minimaxm1scalingtesttimecompute, wang2025stabilizingknowledgepromotingreasoning}. For example, DAPO \cite{yu2025dapoopensourcellmreinforcement} introduces the Clip-Higher strategy, which employs a higher upper clipping threshold to raise the probabilities of low-probability `exploration' tokens, thereby increasing policy entropy. 

However, existing work on policy entropy control is largely confined to understanding the clipping threshold from a static perspective. It lacks a theoretical understanding of the relationship between Gradient-Preserving Clipping and policy entropy control, nor does it tackle entropy control strategy design for effective RL training. To address these limitations, in this paper, we take a Gradient-Preserving Perspective to 
explore the inherent associations of the upper/lower clipping threshold and entropy increase/decrease, and on this basis, aim to flexibly control policy entropy in the training process. Specifically, we focus on two core research questions, laying the foundation of mechanism exploration and strategy design in RLVR:

\begin{rqbox} 
\vspace{-10pt}
  \begin{enumerate}[label=\textbf{RQ\arabic*}, leftmargin=*]
    \item How can we devise the \hltext{regulation} \hltext{mechanism} to precisely control the entropy?
    \item How can we design the \hltext{entropy control strategy} for effective RL training?
  \end{enumerate}
\vspace{-15pt}
\end{rqbox}

The main contributions of our work are three-fold: (1) We theoretically explore the precise contributions of different importance sampling ratio regions to entropy increase and decrease in RL training, and provide the empirical support meanwhile; (2) Based on these insights, we devise the regulation mechanism for flexible entropy control, by applying dynamic modulation to these specific ratio regions; and (3) We leverage this regulation mechanism to further explore diverse entropy control strategies, such as an increase-then-decrease (ID) phase, a decrease-increase-decrease (DID) sequence, and an oscillatory decay (OD) within upper and lower clipping thresholds. Extensive experiments demonstrate the effectiveness of our proposed mechanism and strategy design in mitigating entropy collapse, achieving more precise entropy control and markedly enhancing model performance for RLVR across multiple benchmarks.

\section{Preliminary}
\subsection{RL Algorithms of LLMs}
The PPO-Clip \cite{schulman2017proximal} algorithm addresses the difficulty of determining the appropriate step size in Vanilla Policy Gradient \cite{williams1992simple} methods by clipping the policy update ratio, effectively approximating the trust region constraint of Trust Region Policy Optimization (TRPO) \cite{schulman2015trust}. Specifically, PPO employs an Actor-Critic architecture \cite{konda1999actor}. It relies on limiting the divergence between the new and old policies via clipping, thereby preventing drastic policy drift during single updates.

\begin{figure}[t] 
  \centering
  
  \begin{subfigure}[b]{0.3\textwidth}
    \centering
    \includegraphics[width=\linewidth]{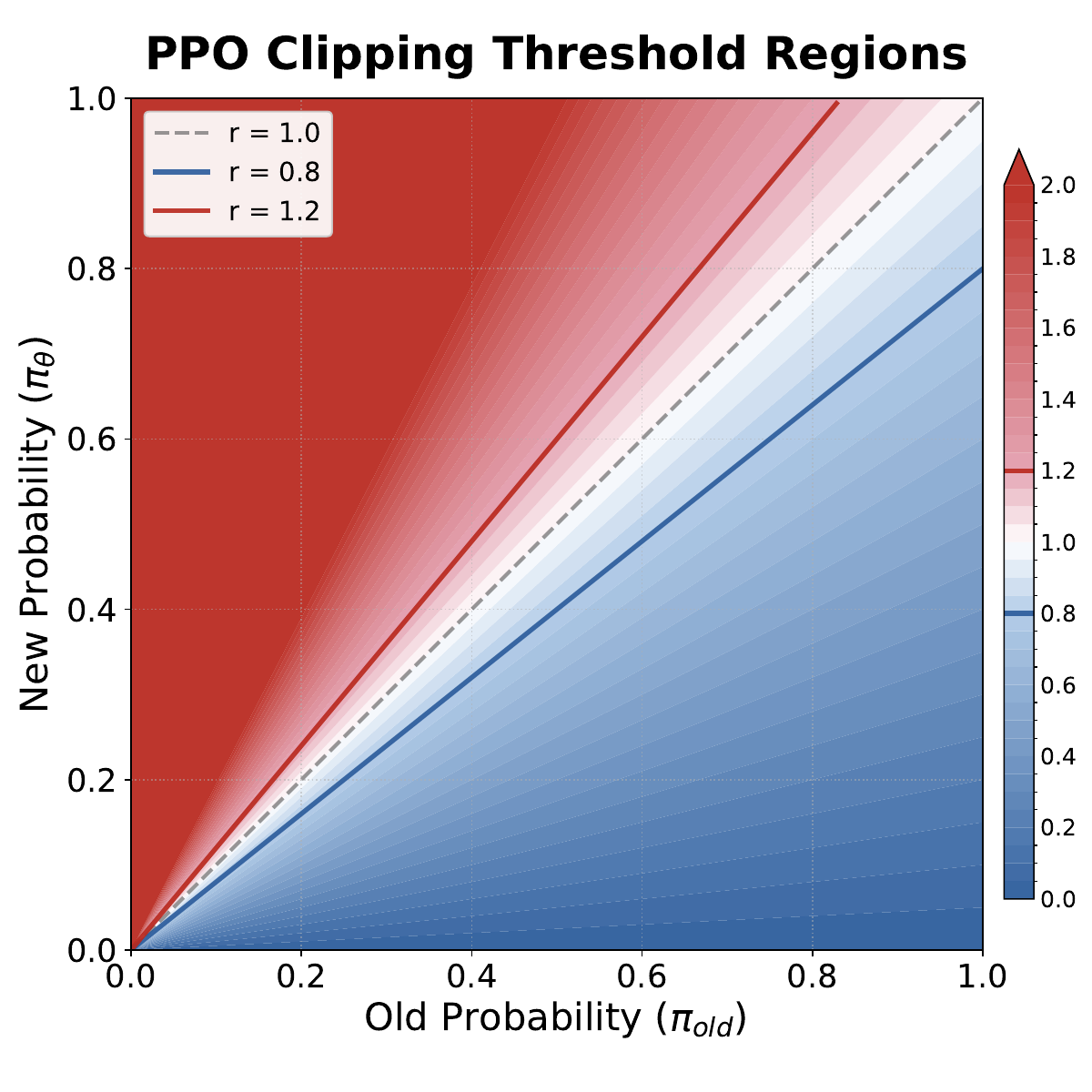} 
    \caption{}
    \label{fig:img1}
  \end{subfigure}
  \begin{subfigure}[b]{0.3\textwidth}
    \centering
    \includegraphics[width=\linewidth]{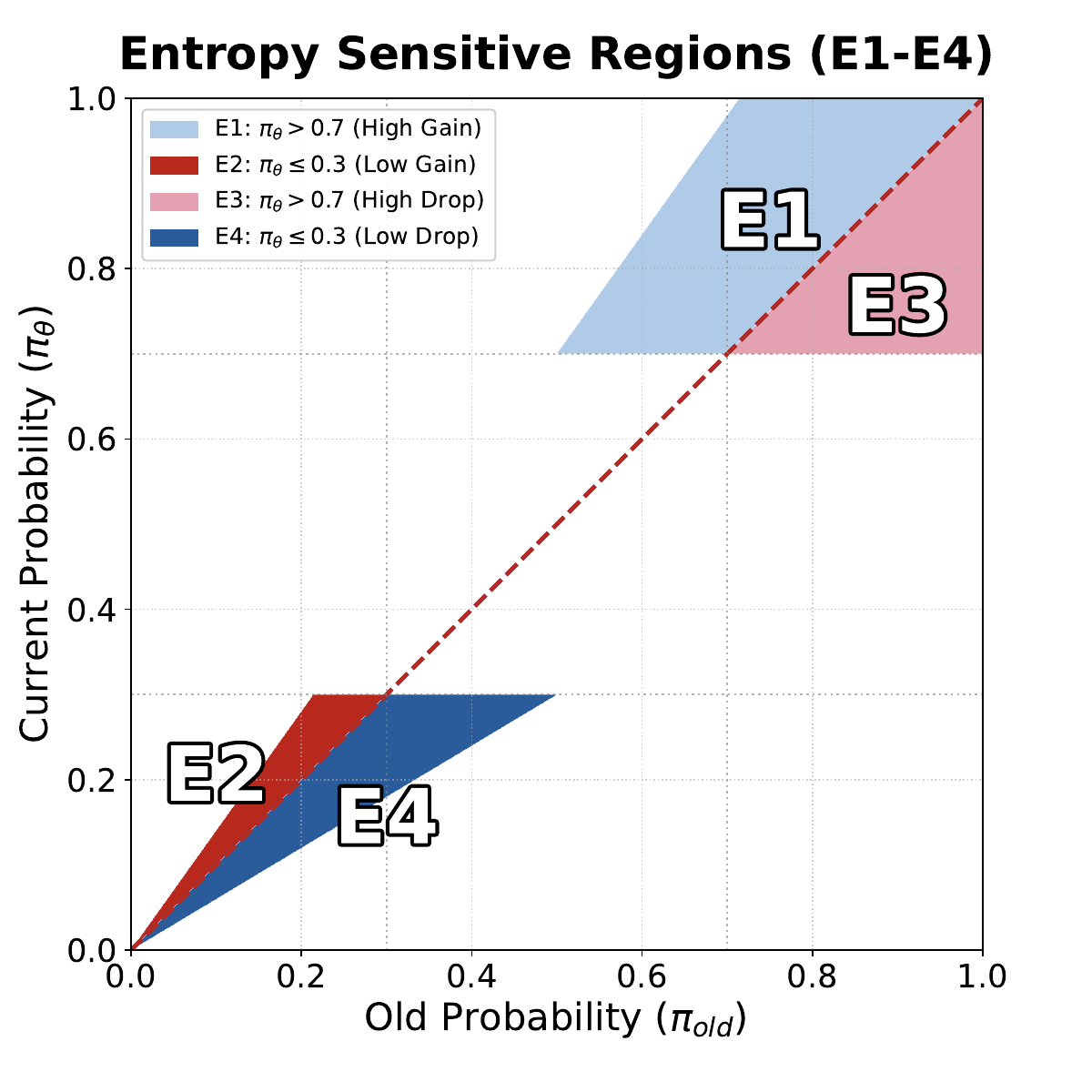}
    \caption{}
    \label{fig:img2}
  \end{subfigure}
    \begin{subfigure}[b]{0.3\textwidth}
    \centering
    \includegraphics[width=\linewidth]{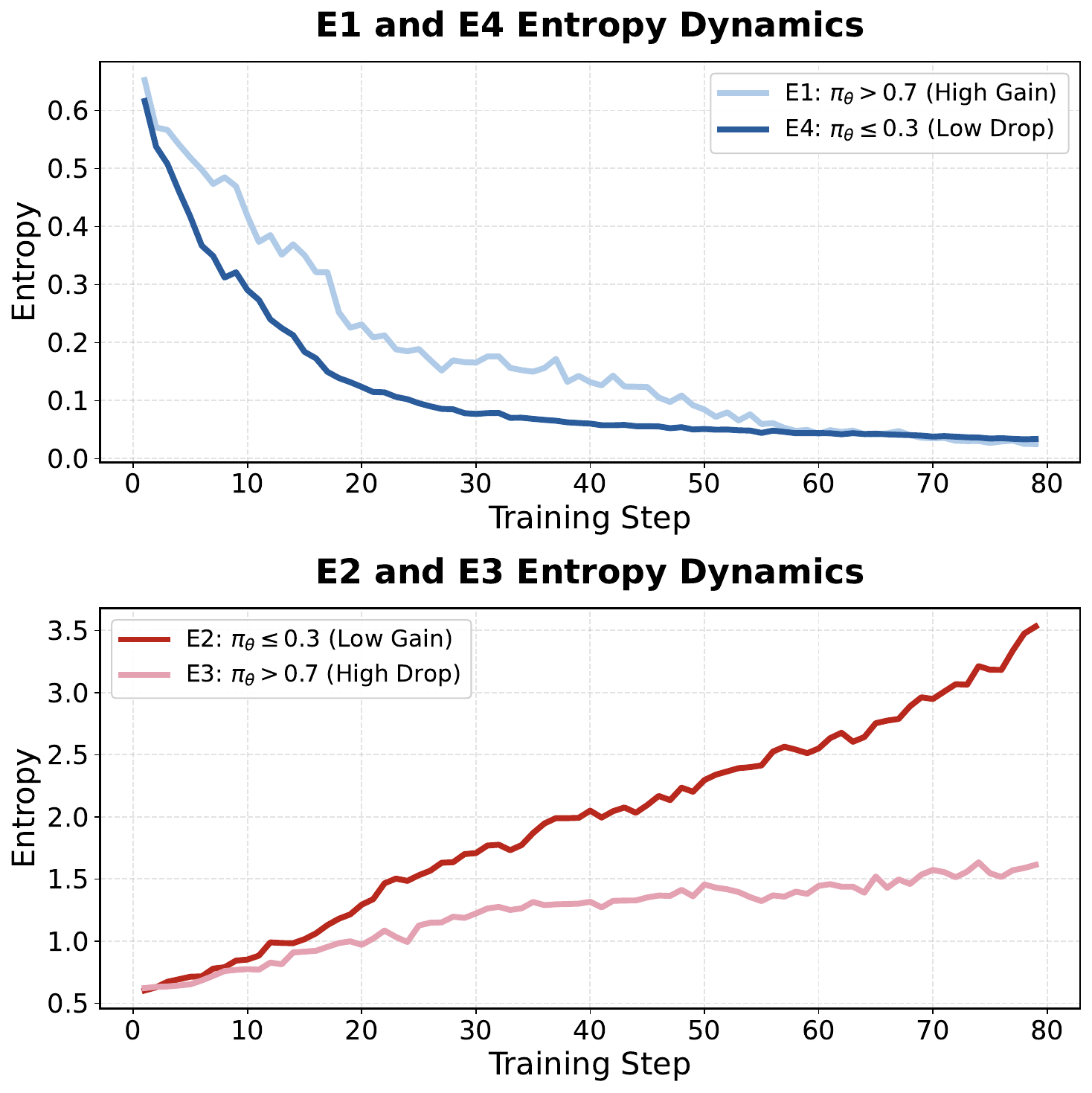}
    \caption{}
    \label{fig:img3}
  \end{subfigure}
\caption{(a) Visualization of PPO clipping threshold regions and probability ratios;  (b) Visualization of four entropy-sensitive regions (E1–E4), categorized by the relationship between the old probability ($\pi_{old}$) and the current probability ($\pi_{\theta}$). These regions distinguish between high ($>0.7$) and low ($\le 0.3$) probability states, as well as probability gains and drops; (c) Entropy dynamics curves showing how regions E1/E4 reduce entropy while E2/E3 increase it.}
\label{fig:three_images}
\end{figure}

The importance sampling ratio $r_t(\theta)=\frac{\pi_{\theta}(a_t|s_t)}{\pi_{old}(a_t|s_t)}$ is defined as the ratio between the new policy $\pi_{\theta}$ and the old policy $\pi_{old}$. The objective function of PPO-Clip, denoted as $L^{CLIP}$, seeks to maximize the advantage function $A_t$ while imposing a constraint on $r_t(\theta)$:
\begin{equation*}
L^{CLIP}(\theta) = \hat{\mathbb{E}}_t \left[ \min \left( r_t(\theta)\hat{A}_t, \text{clip}(r_t(\theta), 1-\epsilon, 1+\epsilon)\hat{A}_t \right) \right]
\end{equation*}
To more conveniently characterize the clipping threshold in PPO-Clip, we adopt a visualization method similar to that in \cite{wang2025stabilizingknowledgepromotingreasoning}, as shown in Figure \ref{fig:img1}. Different colors represent different sizes of the clipping threshold. Generally, we choose $\epsilon_{low}=\epsilon_{high}=0.2$ \cite{schulman2017proximal}.

In PPO-Clip, $\hat{A}_t$ is typically computed using Generalized Advantage Estimation (GAE) \cite{schulman2015high}. To enhance training efficiency and stability, GRPO \cite{shao2024deepseekmath} employs a group-based advantage estimation method that dispenses with the separate Critic. 

\subsection{The Policy Entropy of LLMs}
\label{policy-entropy}
The policy entropy of the model characterizes the degree of uncertainty at the current decision point, or equivalently, the `flatness' of the model's policy. Let $V$ denote the vocabulary space. At time step $t$, given the context $s_t$, the policy entropy $H(\pi_\theta(\cdot|s_t))$ is defined as:
\begin{equation*}
H(\pi_\theta(\cdot|s_t)) = - \sum_{a \in V} \pi_\theta(a|s_t) \log \pi_\theta(a|s_t)
\end{equation*}
The work \cite{cui2025entropymechanismreinforcementlearning} reveals a predictable relationship between policy entropy and performance. Specifically, there exists a negative exponential correlation between the model's policy entropy $H$ and model performance $R$:
\begin{equation*}
R=-a \times exp(H) + b
\end{equation*}
Through theoretical analysis, the work \cite{shen2025entropycontrolllmrlalgorithms} demonstrates that entropy collapse severely impairs the model's output diversity. Consequently, this restricts the update gradients during continual training, leading to a degradation in the model's final performance. Specifically, in RL training without entropy control, the norm of the final policy gradient is limited by the policy entropy:
\begin{equation*}
\Vert \nabla V^{\pi_\theta}(\mathcal{D}) \Vert \leq 2H(\pi_\theta)
\end{equation*}

\section{Theoretical and Empirical Investigations}
\label{methdology1}
To address RQ~1: ``How can we devise the \textbf{regulation} \textbf{mechanism} to precisely control the entropy'', we conduct theoretical analyses regarding the angle between RL training gradients and model entropy gradients. The theoretical  analysis identifies the precise effects of four distinct importance sampling ratio regions on entropy dynamics during training. For empirical investigation, similar to those in \citet{su2025cegppocoordinatingentropygradientpreserving} and \citet{hao2025rethinkingentropyinterventionsrlvr}, we verify the impacts of these regions on policy entropy.

We first provide the theoretical analysis. For a specific token $a$, we define the surrogate objective function for a single-step update as:
\begin{equation}
    L(\theta) = \frac{\pi_\theta(a|s)}{\pi_{\text{old}}(a|s)}\hat{A}
\end{equation}
where $\pi_\theta(a|s)$ denotes the probability of the token under the current policy, $\pi_{\text{old}}(a|s)$ represents the probability under the sampling policy, and $\hat{A}$ is the advantage function. Assuming the policy $\pi_\theta$ is parameterized by logits $z$ (where $\pi(x) = \text{Softmax}(z)_x$), the gradient of the objective function $L$ for a specific token $a$ with respect to the logits $z$ is:
\begin{equation}
    \nabla_z L \propto \hat{A} \cdot \nabla_z \ln \pi_\theta(a|s) = \hat{A} (\mathbf{e}_a - \mathbf{p})
\end{equation}
where $\mathbf{e}_a$ is the one-hot vector for token $a$, and $\mathbf{p}$ is the probability vector over the entire vocabulary.

Next, we consider the global entropy: 
\begin{equation}
    H(\pi)=-\sum_{x \in V} p_x \ln p_x
\end{equation}
The gradient of $H$ with respect to the logits $z$ is derived as:
\begin{equation}
    \nabla_z H = - \mathbf{p} \odot (\ln \mathbf{p} + H \cdot \mathbf{1})
\end{equation}
where $\odot$ denotes element-wise multiplication and $\mathbf{1}$ is a vector of ones. To determine whether the reinforcement learning update increases or decreases entropy, we compute the inner product between the objective gradient $\nabla_z L$ and the global entropy gradient $\nabla_z H$:
\begin{align}
    \langle \nabla_z L, &\nabla_z H \rangle \propto \hat{A} (\mathbf{e}_a - \mathbf{p})^\top \left[ - \mathbf{p} \odot (\ln \mathbf{p} + H \cdot \mathbf{1}) \right] \notag \\
    &= -\hat{A} \left[ \underbrace{p_a(\ln p_a + H)}_{\text{Token-specific term}} - \underbrace{\sum_{x \in V} p_x^2 (\ln p_x + H)}_{\text{Global baseline term}} \right]
    \label{eq1}
\end{align}
For a more detailed derivation, please refer to Section \ref{appendix:proof1}. The second term represents an expectation over the vocabulary (weighted by squared probabilities). Since we are analyzing the gradient contribution of a specific token update, the sign is primarily determined by the token-specific term relative to the baseline. Let
\begin{equation}
    B(\mathbf{p})=\sum_{x \in V} p_x^2(\ln p_x+H).
\end{equation}
Then the local entropy effect of updating token $a$ is governed by:
\begin{equation}
    \operatorname{sgn}(\langle \nabla_z L, \nabla_z H \rangle)
    =
    -\operatorname{sgn}\left(\hat{A}\cdot\left[p_a(\ln p_a+H)-B(\mathbf{p})\right]\right).
\end{equation}

Assuming positive advantage ($\hat{A} > 0$):

\begin{itemize}[left=5pt]
    \item \textbf{Entropy Decreases (E1):} If $p_a(\ln p_a + H) > B(\mathbf{p})$, which is typically satisfied when $-\ln \pi(a|s) < H$, then encouraging it further sharpens the distribution.
    \item \textbf{Entropy Increases (E2):} If $p_a(\ln p_a + H) < B(\mathbf{p})$, which is typically satisfied when $-\ln \pi(a|s) > H$, then  discouraging it flattens the distribution.
\end{itemize}

Assuming negative advantage ($\hat{A} < 0$):

\begin{itemize}[left=5pt]
    \item \textbf{Entropy Increases (E3):} if $p_a(\ln p_a+H) > B(\mathbf{p})$, which is typically satisfied when $-\ln \pi(a|s) < H$, then  discouraging it flattens the distribution.
    \item \textbf{Entropy Increases (E4):} if $p_a(\ln p_a+H) < B(\mathbf{p})$, which is typically satisfied when $-\ln \pi(a|s) > H$, then discouraging it further sharpens the distribution.
\end{itemize}

To validate our theoretical findings, we test the four regions (E1–E4) within the extended PPO-Clip trust region ($0.7 < r < 1.3$) characterized by high ($\pi_{\theta} > 0.7$) or low ($\pi_{\theta} < 0.3$) probability (Figure \ref{fig:img2}). We applied Gradient-Preserving Clipping exclusively to these regions, while ensuring theoretically unaffected tokens continued training normally. Results are shown in Figure \ref{fig:img3}.

\section{Methodology}
Based on the above understanding of how each PPO-Clip region affects entropy, Section \ref{methdology2} builds upon these findings to propose a regulation mechanism capable of stably controlling entropy fluctuations. To address RQ~2: ``How can we design the entropy control strategy for effective RL training?'', we present three training strategies in Section \ref{methdology3} based on rational entropy evolution.

\subsection{Dynamic Clipping Thresholds for Entropy Regulation}
\label{methdology2}

We propose to regulate entropy variations by dynamically adjusting the upper and lower clipping thresholds in PPO-Clip. The key idea is to make the clipping threshold probability-dependent, so that policy updates can be selectively encouraged or constrained according to both the advantage sign and the current token probability.

For positive-advantage tokens ($\hat{A}>0$), the upper clipping threshold controls how much the current policy probability is allowed to increase relative to the rollout policy. DAPO \cite{yu2025dapoopensourcellmreinforcement} argues that a larger upper clipping threshold can facilitate the probability increase of low-probability tokens in positive samples. However, using a uniformly large threshold also allows already high-probability 
\begin{wrapfigure}{r}{0.55\textwidth}
    \centering
    \vspace{-10pt}
    \begin{subfigure}[b]{0.26\textwidth}
        \centering
        \includegraphics[width=\linewidth]{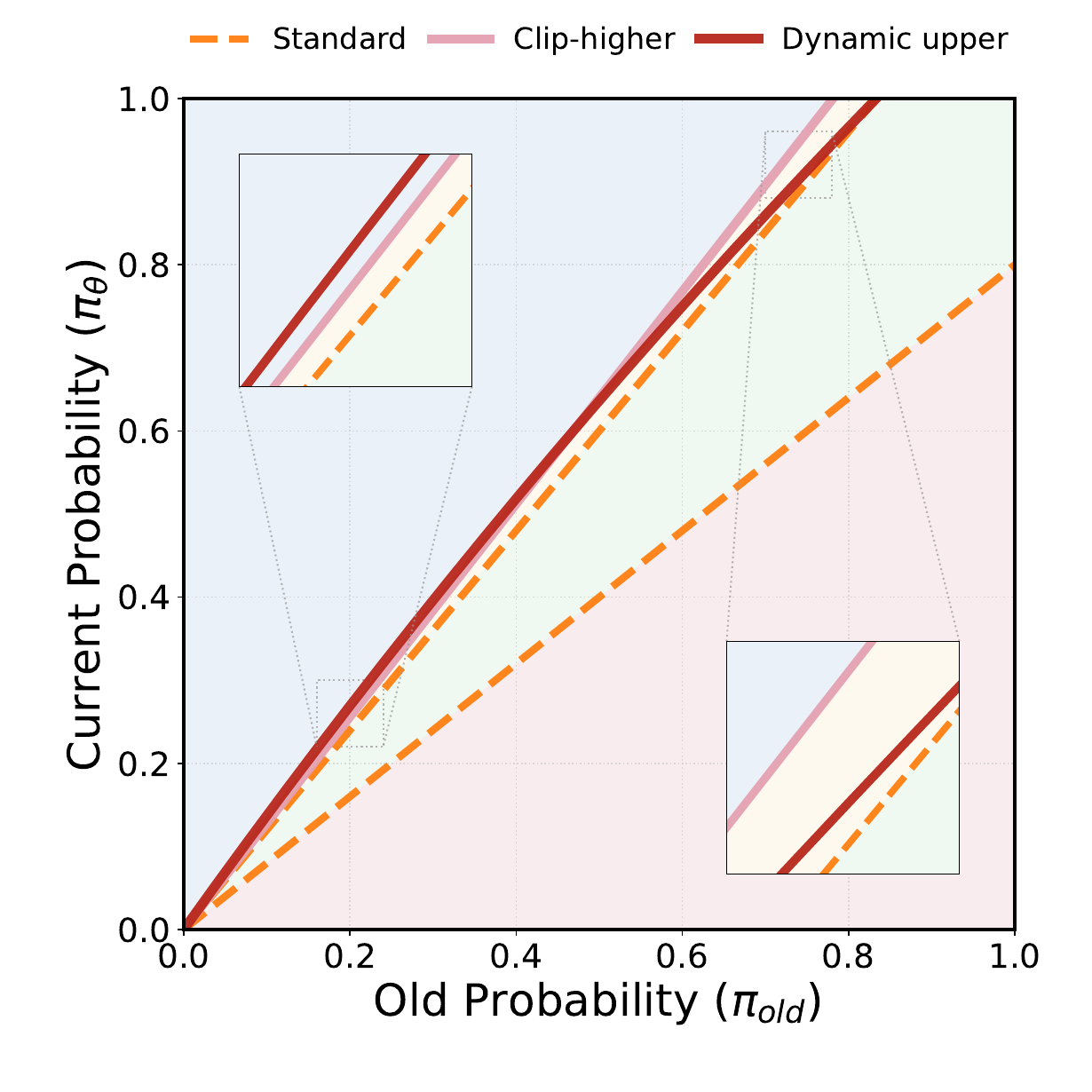}
        \caption{}
        \label{fig:clip-threshold-1}
    \end{subfigure}
    \hfill
    \begin{subfigure}[b]{0.26\textwidth}
        \centering
        \includegraphics[width=\linewidth]{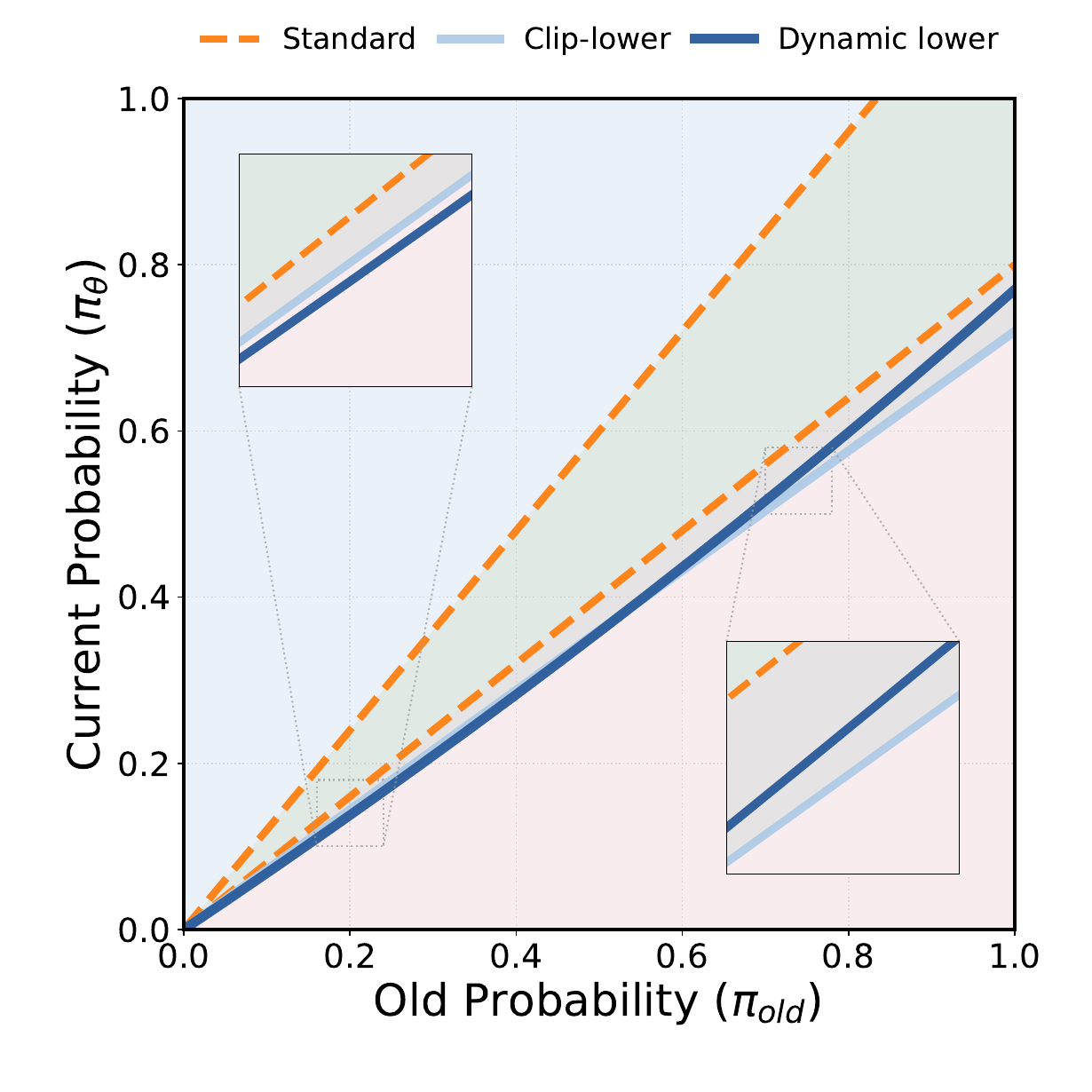}
        \caption{}
        \label{fig:clip-threshold-2}
    \end{subfigure}
    \caption{(a) Dynamic upper clipping threshold; (b) Dynamic lower clipping threshold.}
    \label{fig:clip-threshold}
    \vspace{-10pt}
\end{wrapfigure}
tokens to keep receiving gradients, which may over-concentrate the policy distribution and reduce output diversity. Therefore, we increase the upper clipping threshold for low-probability tokens while decreasing it for high-probability tokens, encouraging exploration without over-optimizing dominant tokens.

For negative-advantage tokens ($\hat{A}<0$), the lower clipping threshold controls how much the current policy probability is allowed to decrease. Unlike the positive-advantage case, negative signals may introduce additional instability \cite{gao2025softadaptivepolicyoptimization}. Let $z$ denote the logits over vocabulary $V$, and let $x$ be a generic token. For a sampled token $a$, the gradient of the log-probability objective with respect to $z_x$ is:
\begin{equation}
    \begin{split}
        \frac{\partial (\ln \pi_\theta(a \mid s) \cdot \hat{A})}{\partial z_x} 
        &= \frac{\partial \pi_\theta(a \mid s)}{\partial z_x} 
        \cdot \frac{\hat{A}}{\pi_\theta(a \mid s)} =
        \begin{cases} 
        (1 - \pi_\theta(a \mid s)) \cdot \hat{A}, & \text{if } x = a, \\
        -\pi_\theta(x \mid s) \cdot \hat{A}, & \text{otherwise}.
        \end{cases}
    \end{split}
    \label{eq:7}
\end{equation}
A detailed analysis of Equation \ref{eq:7} is provided in Section \ref{appendix:proof2}. Since softmax normalization forces the probabilities of other tokens to rise when the sampled token is penalized, inaccurate tokens may be unintentionally boosted. A fixed large lower clipping threshold can further amplify this effect, especially for high-probability negative samples, leading to significant distribution shifts. Thus, we reduce the lower clipping threshold for high-probability negative tokens to preserve update stability. Conversely, for low-probability negative tokens, their influence on the global distribution is limited; moderately relaxing the lower clipping threshold allows them to receive sufficient negative gradients, helping suppress sub-optimal regions and mitigate ineffective clipping.

\begin{figure}[t]
\centering
\begin{subfigure}[b]{0.49\textwidth}
    \centering
    \includegraphics[width=1\linewidth]{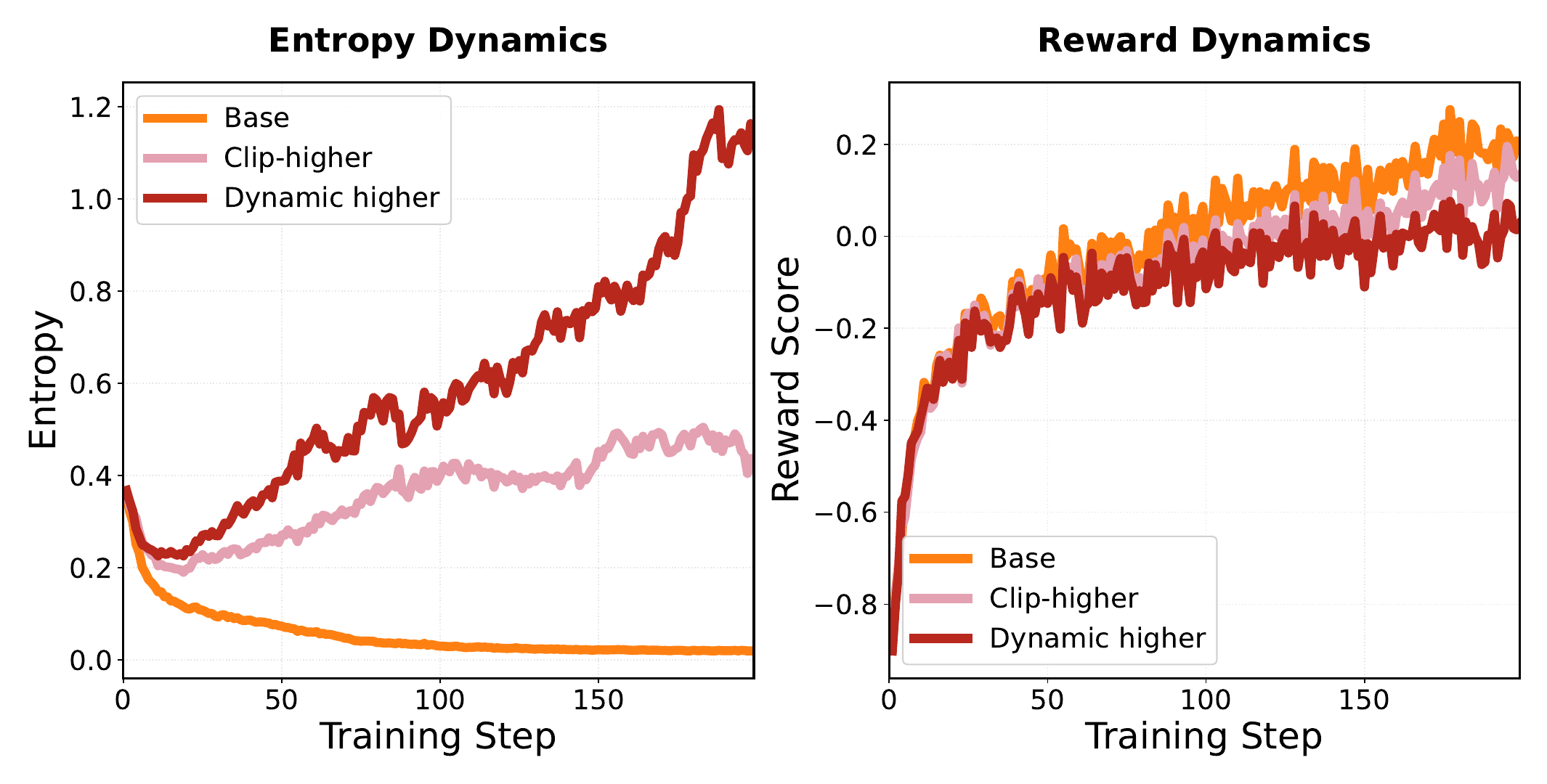} 
    \caption{Dynamic Upper Clipping Threshold}
    \label{upper-side}
\end{subfigure}
\begin{subfigure}[b]{0.49\textwidth}
    \centering
    \includegraphics[width=1\linewidth]{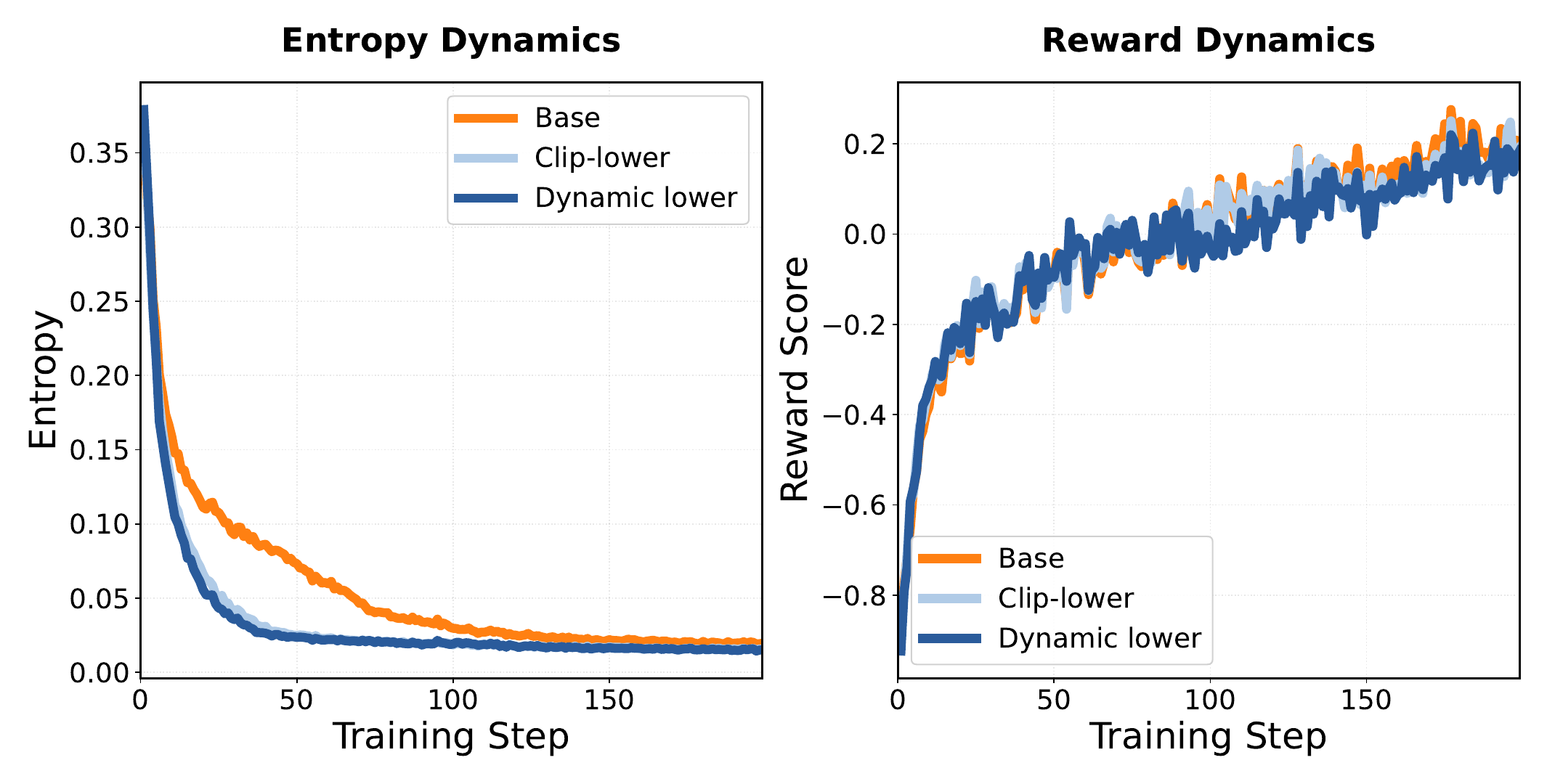} 
    \caption{Dynamic Lower Clipping Threshold}
    \label{lower-side}
\end{subfigure}
\caption{Experimental curves of model entropy regulation. (a) and (b) are training experimental curves with different clipping thresholds.}
\label{fig:6}
\vspace{-10pt}
\end{figure}

To unify both cases, we reformulate the clipping threshold $\epsilon$ as a dynamic function of the current policy probability with stop-gradient:
\begin{equation}
    \epsilon(\pi_{\theta}) := f(\bar{p}_t), \qquad \bar{p}_t = \operatorname{sg}\!\left[\pi_{\theta}(a_t|s_t)\right],
\end{equation}
where $\operatorname{sg}[\cdot]$ denotes a stop-gradient operation. Here, $f(\bar{p}_t)$ is negatively correlated with $\bar{p}_t$. In particular, when adopting a linear form 
$\epsilon(\pi_{\theta}) = \alpha \cdot \bar{p}_t + \beta$, the constraint between the current policy and the rollout policy becomes:
\begin{equation}
    \pi_{\theta}(a_t|s_t) 
    \le 
    \frac{1+\beta}{1-\alpha \cdot\pi_{\theta_{old}}(a_t|s_t)} 
    \cdot \pi_{\theta_{old}}(a_t|s_t)
\end{equation}
As shown in Figure \ref{fig:clip-threshold}, the proposed dynamic thresholds are calibrated to be larger in the low-probability regime and smaller in the high-probability regime. For the upper threshold, this expands updates in the E2 region while constraining E1, thereby increasing entropy by promoting low-probability positive tokens. For the lower threshold, it expands updates in the E4 region while constraining E3, thereby decreasing entropy by suppressing low-probability negative tokens without destabilizing high-probability regions. Consistent with our analysis, Figure \ref{upper-side} shows a steady entropy increase under the dynamic upper threshold, while Figure \ref{lower-side} shows a steady entropy decrease under the dynamic lower threshold.

\subsection{Strategy Design for Entropy Control }
\label{methdology3}

As shown in Section \ref{policy-entropy}, RL training should follow a dynamic entropy strategy: maintaining high entropy in the early stages to promote flexible \textbf{exploration}, while gradually reducing it in later stages to achieve optimal \textbf{performance and output stability}.  Excessive entropy causes instability, while premature reduction hinders exploration. Leveraging the adjustment mechanism from Section \ref{methdology2}, we further design entropy control strategies for LLMs.

\begin{figure}[t]
\begin{center}
    \includegraphics[width=0.95\linewidth]{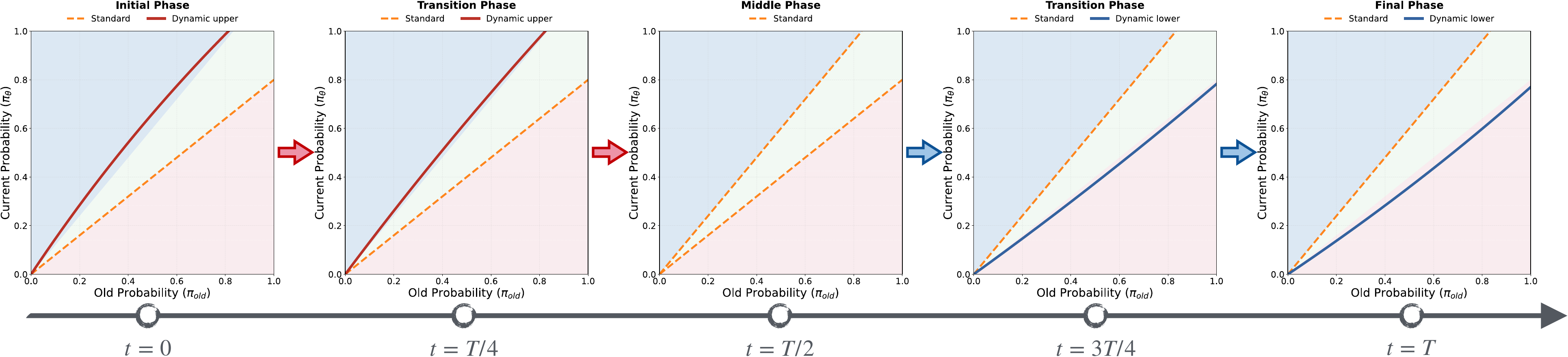}
    \caption{Increase-then-Decrease Entropy control strategy.}
    \label{ID_plot}
\end{center}
    \vspace{-5pt}
\end{figure}
\begin{figure}[t]
\begin{center}
    \includegraphics[width=0.95\linewidth]{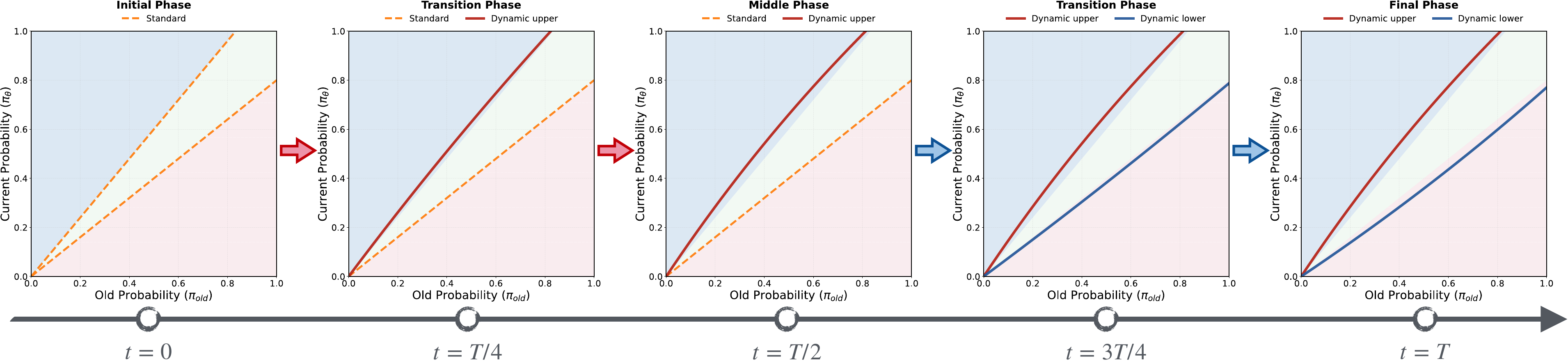}
    \caption{Decrease-Increase-Decrease Entropy control strategy.}
    \label{DID_plot}
    \vspace{-15pt}
\end{center}
\end{figure}

We generalize the PPO clipped objective function into a time-dependent formulation. At training step $k \in [0, T_{max}]$, the loss function is defined as:
\begin{equation*}
\begin{split}
L^{CLIP}_k(\theta) = \hat{\mathbb{E}} \Bigg[ \frac{1}{G} \sum_{i=1}^G \frac{1}{|o_i|}\sum_{t=1}^{|o_i|}  \min \Big( r_{i,t}(\theta) A_i, \text{clip}(r_{i,t}(\theta)&, 1-{\color{red}\mathcal{E}^-_k(p_{i,t})}, 1+{\color{red}\mathcal{E}^+_k(p_{i,t})}) A_i \Big)  \Bigg]
\end{split}
\end{equation*}
The clipping threshold $\color{red}\mathcal{E}_k$ is a function of the detached current policy probability $p_{i,t}=\operatorname{sg}[\pi_\theta(a_{i,t}|s_{i,t})]$, with parameters that evolve over the global step $k$. This function regulates the scaling of the upper and lower bounds according to the current training stage without introducing additional gradients through the clipping thresholds. We explore three specific control strategies: (1) an increase-then-decrease (ID) mode; (2) a decrease-increase-decrease (DID) mode; and (3) an oscillatory decay (OD) mode within a limited entropy bound.

\textbf{Increase-then-Decrease}: We define $\epsilon_{std} = 0.2$ and a temporal scaling factor $\lambda(k) = 1 - \frac{2k}{T_{max}}$.
\begin{itemize}[left=5pt]
    \item \textbf{Phase I ($k < T/2$):} The lower clipping threshold is fixed at $\epsilon_{std}$. Simultaneously, we apply a linear decay to the dynamic upper clipping threshold $\mathcal{H}(p)$, gradually annealing it toward $\epsilon_{std}$.
    \item \textbf{Phase II ($k \ge T/2$):}  The upper clipping threshold is fixed at $\epsilon_{std}$. A linear gain progressively transitions the lower threshold from $\epsilon_{std}$ to the dynamic $\mathcal{M}(p)$.
\end{itemize}
The schematic illustration of this scheduling process is presented in Figure \ref{ID_plot}. The resulting training objective function is formulated as follows:
{
$$\begin{aligned}
\mathcal{E}&^+_{k-ID}(p),\mathcal{E}^-_{k-ID}(p) = 
\begin{cases} 
\lambda_k \cdot \mathcal{H}(p) + (1-\lambda_k) \cdot \epsilon_{std}, \quad \epsilon_{std} & 0 \le k \le \frac{T}{2} 
\\
\epsilon_{std},\quad (1+\lambda_k)\cdot \epsilon_{std} -\lambda_k\cdot \mathcal{M}(p)  & \frac{T}{2} < k \le T
\end{cases} 
\end{aligned}$$
}
\textbf{Decrease-Increase-Decrease}: Unlike the ID, the DID control strategy allows the model entropy to first decrease in the first phase, controls the increase of model entropy through gradient clipping before entropy collapse, and then controls model convergence in the second phase.
\begin{itemize}[left=5pt]
    \item \textbf{Phase I ($k < T/2$):} The lower clipping threshold is fixed at $\epsilon_{std}$. A linear gain progressively transitions the upper threshold from $\epsilon_{std}$ to the dynamic $\mathcal{H}(p)$.
    \item \textbf{Phase II ($k \ge T/2$):}  The upper clipping threshold is fixed at $\mathcal{H}(p)$. A linear gain progressively transitions the lower threshold from $\epsilon_{std}$ to the dynamic $\mathcal{M}(p)$.
\end{itemize}
The schematic illustration of this scheduling process is presented in Figure \ref{DID_plot}. The resulting training objective function is formulated as follows:
{
$$\begin{aligned}
\mathcal{E}&^+_{k-DID}(p),\mathcal{E}^-_{k-DID}(p) = 
\begin{cases} 
\lambda_k \cdot \epsilon_{std} + (1-\lambda_k) \cdot \mathcal{H}(p), \quad \epsilon_{std} & 0 \le k \le \frac{T}{2} 
\\
\mathcal{H}(p),\quad (1+\lambda_k)\cdot \epsilon_{std} -\lambda_k\cdot \mathcal{M}(p) & \frac{T}{2} < k \le T
\end{cases} 
\end{aligned}$$
}
\textbf{Oscillatory Decay:} Both ID and DID  control strategies partition the training process into two distinct stages. In contrast, the OD  control strategy is designed to enable the model to undergo autonomous oscillatory attenuation throughout the training duration. Specifically, we begin by defining a pair of entropy thresholds that evolve in relation to the training step $k$:
\begin{equation*}
\begin{aligned}
\tau_{low} &= H_{min}\textrm{ ,} \qquad \tau_{high}(k) = H_{min} + (H_{init} - H_{min}) \cdot (1 - \frac{k}{T})
\end{aligned}
\end{equation*}
where $H(\pi_{\theta_k})$ denotes the entropy of the current policy at training step $k$ and $H_{min}$ represents the target entropy lower bound, which is defined here as $0.2H_{init}$. We introduce a discrete state variable $s_k \in \{0, 1\}$ to characterize the current control mode (where $1$ signifies the entropy-increasing mode and $0$ signifies the entropy-decreasing mode), initialized as $s_0=0$. 
$$s_k = \begin{cases} 
1 & \text{if } H(\pi_{\theta_k}) \le \tau_{low} \quad (\text{Trigger Boost}) \\
0 & \text{if } H(\pi_{\theta_k}) > \tau_{high}(k) \quad (\text{Trigger Suppress}) \\
s_{k-1} & \text{otherwise} \quad (\text{Hold})
\end{cases}$$
Based on the current state $s_k$, the dynamic clipping threshold $\mathcal{E}^+_k(p)$ and $\mathcal{E}^-_k(p)$ are defined as follows:
$$\begin{aligned}
\mathcal{E}^+_{k-OD}(p), \mathcal{E}^-_{k-OD}(p) = 
&\begin{cases} 
\mathcal{H}(p), \quad \epsilon_{std} & \text{if } s_k = 1 
\\
\epsilon_{std}, \quad \mathcal{M}(p) & \text{if } s_k = 0\textrm{ .}
\end{cases} 
\end{aligned}$$

\section{Experiments}

\subsection{Experimental Setup}
To validate our proposed training strategies, we train Qwen2.5-Math-7B \cite{yang2024qwen25mathtechnicalreportmathematical} and Qwen2.5-7B \cite{yang2024qwen25mathtechnicalreportmathematical} on the DAPO-MATH dataset. We conducted a comprehensive evaluation of mathematical performance across the AIME24 \cite{aime24}, AIME25 \cite{aime25}, GSM8k \cite{cobbe2021trainingverifierssolvemath}, AMC, MATH-500, and Olympiad \cite{lightman2023let} benchmarks. To ensure reliability, we sample responses per problem: 32 for AIME24, AIME25, and AMC; 4 for MATH-500 and Olympiad; and 2 for GSM8k. 

In addition to the original GRPO, we selected several baselines, including Clip-Higher \cite{yu2025dapoopensourcellmreinforcement}, Clip-Lower, Entropy-Regularization, Clip-Cov \cite{cui2025entropymechanismreinforcementlearning}, GSPO \cite{zheng2025groupsequencepolicyoptimization}, and SAPO \cite{gao2025softadaptivepolicyoptimization}. To further examine the generality of our method, we additionally evaluate non-mathematical capabilities on GPQA-Diamond \cite{rein2023gpqa}, LiveCodeBench \cite{jain2024livecodebench}, and MMLU-Redux \cite{gema2024mmlu} using Qwen2.5-Math-7B, and conduct experiments on Phi-4-14B \cite{abdin2024phi}, a non-Qwen model family, trained for 200 steps.

Training configurations included a learning rate of $1\times10^{-6}$, a sampling rate of 8 responses per prompt, and a global batch size of 512. The maximum response length was set to 4096 tokens for Qwen2.5-Math-7B and 8192 tokens for Qwen2.5-7B. Further training details are provided in Section \ref{Appendix:exp-setup}. Further evaluation details are provided in Section \ref{Appendix:eval-setup}.

\begin{figure}[t]
\begin{center}
    \includegraphics[width=1\linewidth]{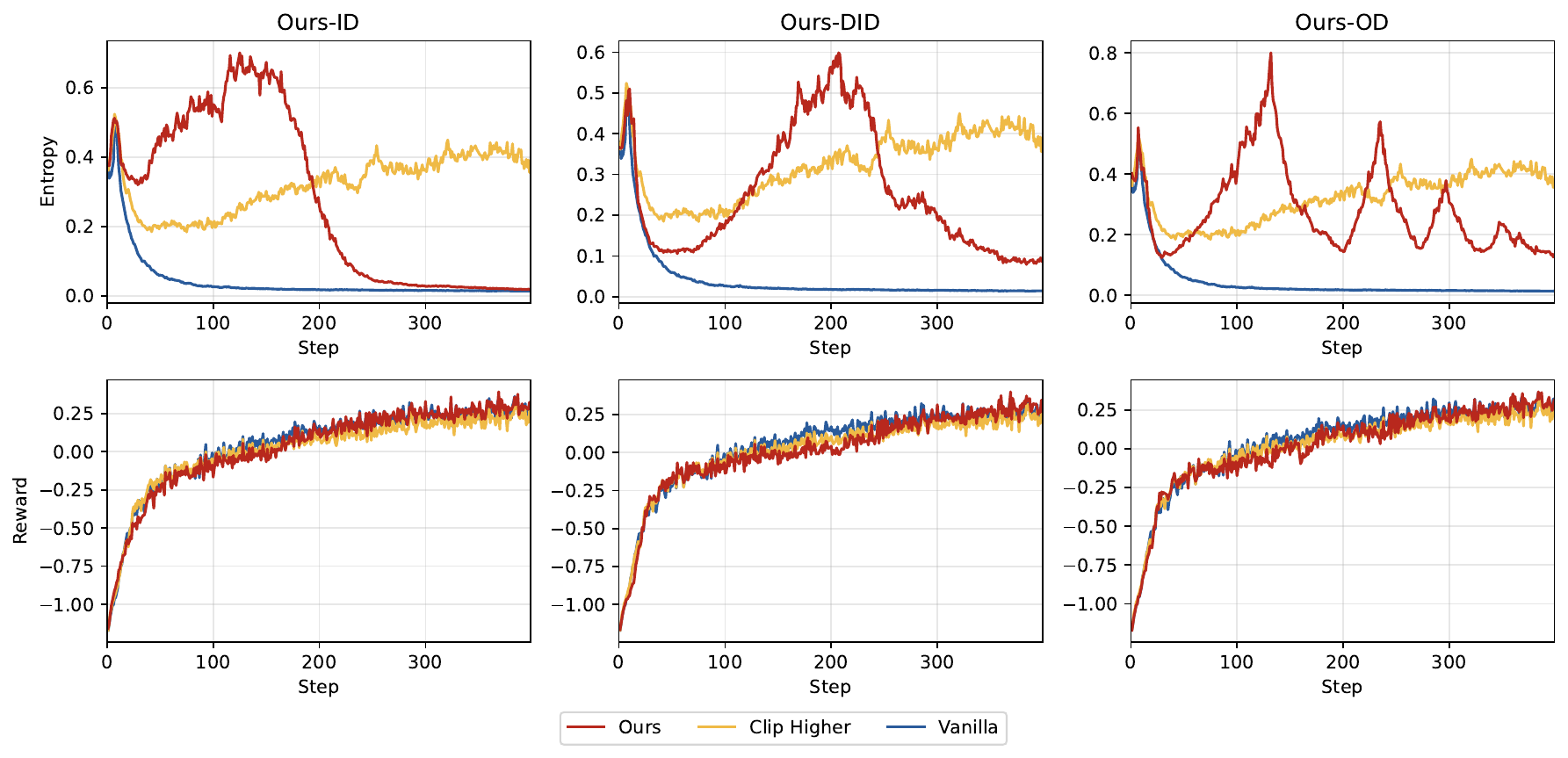}
    \vspace{-15pt}
    \caption{Curves showing changes in Entropy and Reward during the training process of Qwen2.5-Math-7B for various training methods }
    \label{fig:analysis1}
    \vspace{-10pt}
\end{center}
\end{figure}

\subsection{Experimental Results and Analysis}
The results of the training experiments on the dynamic upper clipping threshold and dynamic lower clipping threshold are shown in Figure \ref{fig:6}, which demonstrates that our control over entropy increase and entropy decrease is effective.

The experimental results are shown in Table \ref{tab:1}. We can see that our three training strategies consistently improve performance across multiple mathematical benchmarks. To further verify the generality of our method, we additionally evaluate its effectiveness on non-math domains and on a different model family. As shown in the left panel of Table \ref{tab:addition}, our method achieves consistent gains on non-math benchmarks. Moreover, the right panel of Table \ref{tab:addition} reports results on Phi-4-14B, where our strategies also outperform GRPO on most benchmarks, demonstrating scalability to larger models and generalization beyond the Qwen model family.

\begin{table*}[h]
\centering
\caption{Performance on mathematical task benchmarks.}
\label{tab:1}
\vspace{-6pt}
\scriptsize
\setlength{\tabcolsep}{2.0pt}
\renewcommand{\arraystretch}{0.92}

\resizebox{\textwidth}{!}{%
\begin{tabular}{lcccccc@{\hspace{8pt}}cccccc}
\toprule
\multirow{2}{*}{\textbf{Method}}
& \multicolumn{6}{c}{\textbf{Qwen2.5-MATH-7B}}
& \multicolumn{6}{c}{\textbf{Qwen2.5-7B}} \\
\cmidrule(lr){2-7}\cmidrule(lr){8-13}
& \textbf{AIME24} & \textbf{AIME25} & \textbf{AMC} & \textbf{MATH} & \textbf{GSM8K} & \textbf{Olymp.}
& \textbf{AIME24} & \textbf{AIME25} & \textbf{AMC} & \textbf{MATH} & \textbf{GSM8K} & \textbf{Olymp.} \\
\midrule

GRPO
& 24.4 & 10.2 & 54.8 & 81.1 & 91.6 & 46.2
& 15.9 & 9.6  & 44.9 & 77.1 & 90.5 & 41.9 \\

Clip-H
& 30.5 & 12.7 & 55.8 & 82.7 & \underline{93.3} & 46.4
& 16.5 & 11.6 & 46.6 & \underline{77.6} & 91.6 & 42.6 \\

Clip-L
& 22.9 & 8.1  & 53.2 & 80.8 & 79.9 & 44.0
& 16.4 & 8.5  & 42.2 & 77.0 & 89.9 & 40.0 \\

Ent-Reg
& 21.6 & 12.7 & 53.0 & 79.6 & 91.2 & 44.2
& 16.2 & 11.3 & 43.4 & 76.3 & 91.5 & 40.5 \\

Clip-Cov
& 27.1 & 17.1 & 53.9 & \underline{83.6} & 92.2 & 44.3
& 13.9 & 9.0  & 44.1 & 77.3 & 89.3 & 42.2 \\

SAPO
& 31.1 & 10.8 & 56.6 & 82.3 & 91.7 & 45.1
& 13.1 & 8.9  & 42.0 & 74.9 & 84.1 & 42.0 \\

GSPO
& 27.9 & 10.4 & 57.3 & 81.9 & 85.5 & 40.5
& 16.4 & 9.1  & \underline{47.4} & 77.2 & \underline{92.4} & 42.1 \\

\rowcolor{TableBlue}
Ours-ID
& \textbf{33.1} & \underline{18.1} & \textbf{58.9} & \textbf{84.5} & \textbf{93.7} & \textbf{48.3}
& \textbf{17.0} & 9.3 & 45.3 & \textbf{78.0} & 92.2 & 42.6 \\

\rowcolor{TableBlue}
Ours-DID
& \underline{31.5} & 15.2 & 55.0 & 82.3 & 91.7 & \underline{47.8}
& \underline{16.9} & \textbf{12.1} & 45.3 & 77.5 & \textbf{93.4} & 42.4 \\

\rowcolor{TableBlue}
Ours-OD
& 31.1 & \textbf{18.3} & \underline{58.3} & 83.0 & 92.4 & 47.5
& \underline{16.9} & \underline{11.7} & \textbf{48.3} & 77.2 & 92.0 & \textbf{44.0} \\

\bottomrule
\end{tabular}%
}
\vspace{-8pt}
\end{table*}

\begin{table*}[h]
\centering
\small
\caption{Additional evaluation on non-math domains and Phi-4-14B. Left: results on non-math benchmarks using Qwen2.5-Math-7B. Right: results on Phi-4-14B trained for 200 steps.}
\label{tab:addition}

\begin{minipage}[t]{0.52\textwidth}
\centering
\resizebox{\linewidth}{!}{
\begin{tabular}{lccc}
\toprule
Method & GPQA-Diamond & LiveCodeBench & MMLU-Redux \\
\midrule
GRPO
& 37.22 & 15.83 & 60.18 \\
Ent-Reg
& 29.58 & 17.25 & 59.40 \\
GSPO
& 30.87 & 19.05 & 61.68 \\
SAPO
& 26.45 & 20.28 & 61.96 \\
Clip-Cov
& 27.40 & 19.81 & 62.00 \\
\midrule
\rowcolor{TableBlue}
Ours-ID
& \textbf{38.35} & \textbf{24.83} & 65.72 \\
\rowcolor{TableBlue}
Ours-DID
& 35.83 & 22.09 & 65.16 \\
\rowcolor{TableBlue}
Ours-OD
& 36.84 & 23.13 & \textbf{67.74} \\
\bottomrule
\end{tabular}
}
\end{minipage}
\hfill
\begin{minipage}[t]{0.45\textwidth}
\centering
{\renewcommand{\arraystretch}{1.35}
\resizebox{\linewidth}{!}{
\begin{tabular}{
l
c
>{\columncolor{TableBlue}}c
>{\columncolor{TableBlue}}c
>{\columncolor{TableBlue}}c
}
\toprule
Benchmark & GRPO & Ours-ID & Ours-DID & Ours-OD \\
\midrule
AIME24
& 16.35 & 20.21 & \textbf{23.23} & 20.10 \\
AIME25
& 17.09 & \textbf{24.06} & 17.88 & 17.72 \\
AMC
& 58.90 & 63.25 & \textbf{64.20} & 59.71 \\
MATH-500
& 80.95 & \textbf{85.35} & 84.90 & 81.45 \\
GSM8K
& 94.26 & \textbf{95.45} & 94.81 & 95.15 \\
Olympiad
& 48.15 & \textbf{50.11} & 46.55 & 49.26 \\
\bottomrule
\end{tabular}
}
}
\end{minipage}

\end{table*}

The three strategies target different training regimes. 
\textbf{ID} is best suited for SFT-aligned models, as the initial entropy increase encourages diverse reasoning before convergence, yielding the strongest overall results on Qwen2.5-MATH-7B and strong non-math performance. 
\textbf{DID} is preferable for base or high-entropy models, where early entropy reduction stabilizes optimization; this is reflected in its competitive results on Qwen2.5-7B. 
\textbf{OD} is more suitable for long training or potential local minima, since periodic entropy oscillation maintains exploration and leads to strong competition-level and out-of-domain performance.

We further analyze the entropy and performance (Section \ref{analysis1}), the setting of training phase ratios and clipping function (Section \ref{analysis3}) in the training process of our method. The remaining results including the analysis of the average clipping threshold, the analysis of the clipping probability and the experiment on replacing dynamic clipping threshold with Clip-Higher and Clip-Lower can be found in Section \ref{appendix:experimental}.

\subsection{Analysis of Entropy and Performance}
\label{analysis1}




In Figure \ref{fig:analysis1}, we present the entropy change curves and reward curves during the training process of our three methods compared with the two baselines, GRPO and Clip-Higher, on Qwen2.5-Math-7B. We can analyze some interesting observations:

\begin{wrapfigure}{r}{0.5\textwidth}
    \centering
    \vspace{-10pt}
    \includegraphics[width=\linewidth]{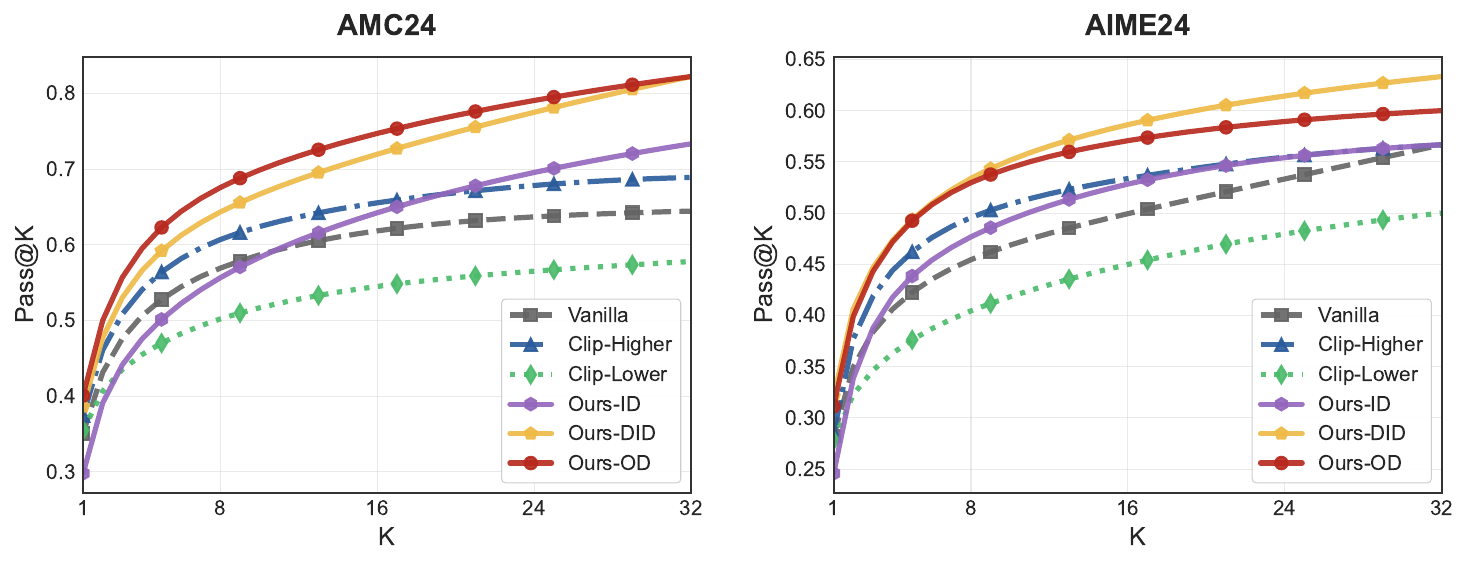}
    \caption{Comparison of Pass@K metrics across various methods.}
    \label{fig:passk}
    \vspace{-10pt}
\end{wrapfigure}

First, our entropy regulation mechanism is effective. By adjusting the clipping threshold during the training process, the change in the model's entropy is clear. Second, our control strategy is effective. The model training reward is low in the early stages of training but surpasses other methods in the later stages. To evaluate the exploration performance of the model in the early stages of training, we show the Pass@32 performance of various methods at the mid-training stage (120 steps) on AMC24 and AIME24 in Figure \ref{fig:passk}. It can be seen that the performance of each method on Pass@1 is similar. However, after increasing the number of outputs, our method has better Pass@K performance.

\subsection{Analysis of Phase Ratios and Clipping Function}
\label{analysis3}

\begin{wrapfigure}{r}{0.5\textwidth}
    \centering
    \vspace{-10pt}
    \includegraphics[width=\linewidth]{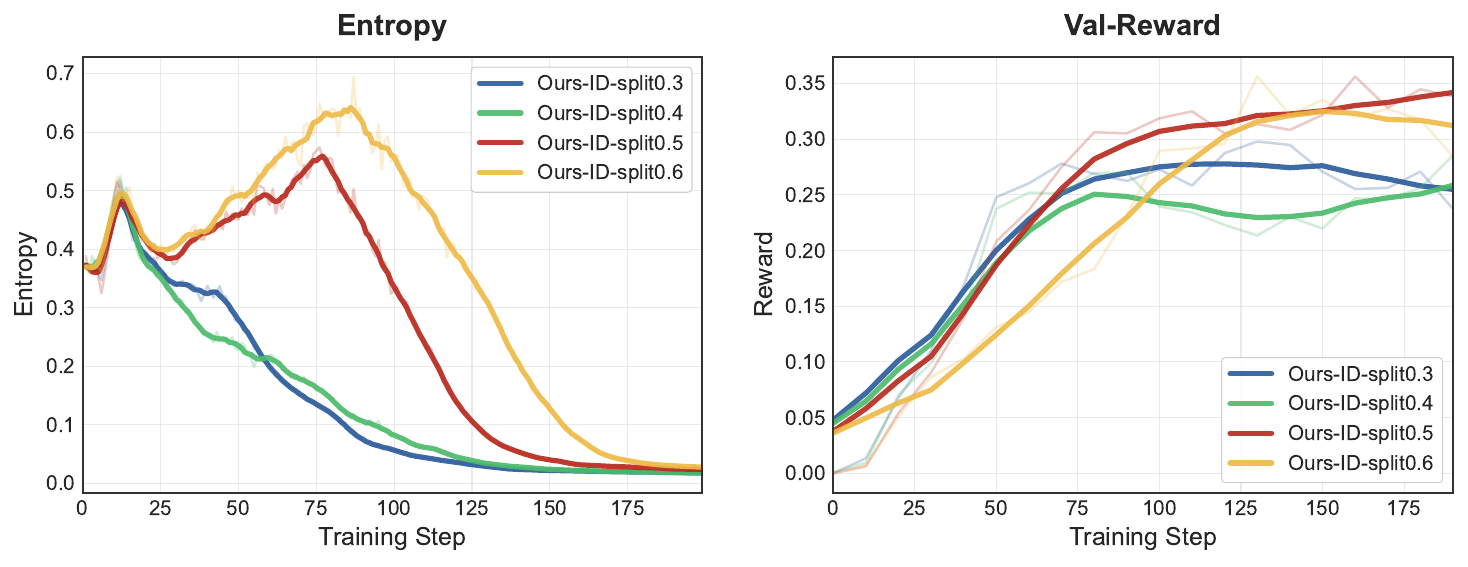}
    \caption{Comparison of entropy and validation set score curves under different phase ratios.}
    \label{fig:time_radio}
    \vspace{-10pt}
\end{wrapfigure}

For Ours-ID and Ours-DID, we split training into an entropy-control phase and a performance-refinement phase. To study the effect of this phase ratio, we vary the length of the first phase among $0.3$, $0.4$, $0.5$, and $0.6$, while keeping all other settings unchanged, and train Qwen2.5-Math-7B for $200$ steps. The results are shown in Figure~\ref{fig:time_radio}.

Theoretically, the rate of entropy increase (defined here as the acceleration of entropy growth) diminishes during the first stage. If this initial control period is too short, the model’s entropy may begin to decline before reaching a sufficient peak. This explains the entropy trends observed when the stage ratio is set to $0.3$ or $0.4$. Conversely, while a stage ratio of $0.6$ allows the model to attain higher entropy in the first stage, it leads to overly rapid convergence in the second stage. Consequently, as evidenced by the validation accuracy in the right panel of Figure \ref{fig:time_radio}, a stage ratio of $0.5$ yields the optimal performance.

We also analyze the choice of the probability-dependent clipping function. The linear form is adopted for its monotonicity, numerical stability, and computational simplicity; additional ablations in Section \ref{clipping-function} show that it consistently outperforms or matches the exponential variant, supporting our design choice.

\section{Conclusions}
 In this paper, we conduct an in-depth study on dynamic entropy control in RLVR from the perspective of gradient preservation. To address the issue of entropy collapse encountered during GRPO training, we focus on two research issues: (1) regulation mechanism for precisely controlling  entropy variations, and (2) entropy control strategy in RLVR training process. We introduce three entropy control strategies for the training phase including increase-then-decrease, decrease-increase-decrease and oscillatory decay. Extensive experimental performance evaluations and analyses of training curves validate the effectiveness of our method.

\newpage
\bibliographystyle{unsrtnat}
\bibliography{main}

\newpage
\appendix
\startcontents[appendices]
\section*{Appendix Contents}
\printcontents[appendices]{l}{1}{\setcounter{tocdepth}{2}}
\newpage
\section{Related Work}
\subsection{Reinforcement Learning and Entropy in Large Language Models}

Inspired by DeepSeek-R1 \cite{guo2025deepseek}, RLVR has been extensively adopted in the post-training of LLMs, yielding a series of notable contributions \cite{wen2025reinforcementlearningverifiablerewards, huang2025visionr1incentivizingreasoningcapability, cheng2025agentr1trainingpowerfulllm, chen2025metisspecsdecouplingmultimodallearning}. Despite the remarkable success of RLVR, a growing body of literature identifies entropy collapse as a critical challenge within this paradigm \cite{cui2025entropymechanismreinforcementlearning, shen2025entropycontrolllmrlalgorithms}. Specifically, \cite{cui2025entropymechanismreinforcementlearning} analyzes policy entropy collapse as a major obstacle in scaling RL for LLM reasoning and proposes mechanisms to overcome it, while further elucidating the relationship between policy entropy and model performance. Meanwhile, \cite{wang20258020rulehighentropyminority} demonstrate that performing policy gradient updates exclusively on high-entropy tokens can more efficiently enhance model performance—an effect that is particularly pronounced in larger models. Furthermore, through theoretical analysis, \cite{shen2025entropycontrolllmrlalgorithms} establish that entropy collapse severely compromises output diversity; this restricts the gradients available for continuous training, ultimately leading to a degradation in final performance. Finally, \cite{jin2026revisitingentropyreinforcementlearning} delineate the key factors governing entropy dynamics, including the clipping threshold, the number of offline updates, and the diversity of training data. 

\subsection{Control of Entropy in Large Language Models}
Entropy is often a critical metric in RL for LLMs. To mitigate the phenomenon of entropy collapse during the RL process, numerous studies have optimized and improved the framework across multiple dimensions. In traditional approaches \cite{schulman2017proximal, ziebart2008maximum}, entropy maximization is employed by introducing an entropy regularization term into the loss function to prevent a continuous decline in entropy during training. Specifically, \cite{schulman2017proximal} incorporated entropy regularization into the PPO algorithm to maintain policy exploration. However, literature suggests that this regularization method is not consistently effective in large language models; \cite{shen2025entropycontrolllmrlalgorithms} elucidates the reasons behind this inefficacy.

In the context of preventing entropy collapse during model training, DAPO \cite{yu2025dapoopensourcellmreinforcement} was the first to propose the ``Clip-Higher'' strategy. They argue that the original upper clipping threshold truncation restricts the probability increase of low-probability tokens to some extent, thereby limiting the diversity of model generation. Consequently, they employ a larger upper clipping threshold for the importance ratio to avoid clipping low-probability tokens, resulting in a degree of entropy increase. TDPO \cite{wang2025stabilizingknowledgepromotingreasoning} differentiates the clipping range at the token dimension; building on the DAPO approach, it restricts the upper clipping threshold for high-probability tokens in positive samples while extending the lower clipping threshold for low-probability tokens. CISPO \cite{minimax2025minimaxm1scalingtesttimecompute} does not directly mask tokens during the clipping process but retains their gradients to ensure continued participation in training, uniformly clamping only the out-of-range importance weights to predefined upper and lower bounds. Clip-Cov \cite{cui2025entropymechanismreinforcementlearning} restricts token updates by randomly selecting a small subset of tokens with high covariance and either detaching their gradients or applying a KL penalty during the policy gradient update. SAPO \cite{gao2025softadaptivepolicyoptimization} replaces hard clipping with a temperature-controlled smooth gating mechanism to construct a continuous trust region. Although these works attempt to control entropy by manipulating the clipping threshold, they lack a systematic understanding of how the clipping threshold regulates entropy and exhibit limited flexibility.

\section{Theoretical Proofs}
Here, we mainly conduct detailed proofs and derivations for the two conclusions \cref{eq1} and \cref{eq:7} in the paper.
\subsection{Proof of \cref{eq1}}
\label{appendix:proof1}
\textbf{\cref{eq1}:}  The inner product between the objective gradient $\nabla_z L$ and the global entropy gradient $\nabla_z H$ is:
\vspace{-10pt}
\begin{align*}
    \langle \nabla_z L, &\nabla_z H \rangle \propto \hat{A} (\mathbf{e}_a - \mathbf{p})^\top \left[ - \mathbf{p} \odot (\ln \mathbf{p} + H \cdot \mathbf{1}) \right] \notag \\
    &= -\hat{A} \left[ \underbrace{p_a(\ln p_a + H)}_{\text{Token-specific term}} - \underbrace{\sum_{x \in V} p_x^2 (\ln p_x + H)}_{\text{Global baseline term}} \right]
    \label{eq1}
\end{align*}
\textit{Proof.} Let $V$ be the vocabulary set. For a given state, let $\mathbf{z} \in \mathbb{R}^{|V|}$ denote the logits output by the network. The policy distribution $\mathbf{p} = \text{softmax}(\mathbf{z})$ is defined such that for any token $x \in V$:
$$p_x = \frac{e^{z_x}}{\sum_{k \in V} e^{z_k}}$$
The entropy of the policy is defined as:
$$H(\mathbf{p}) = - \sum_{x \in V} p_x \ln p_x$$
To compute gradients with respect to logits $z$, we first establish the partial derivative of the probability $p_x$ with respect to the logit $z_y$:
$$\frac{\partial p_x}{\partial z_y} = p_x (\delta_{xy} - p_y) = \begin{cases} 
p_x (1 - p_x) & \text{if } x = y \\
-p_x p_y & \text{if } x \neq y 
\end{cases}$$
where $\delta_{xy}$ is the Kronecker delta.
We apply the chain rule to find the gradient of entropy $H$ with respect to a specific logit $z_y$:
$$\frac{\partial H}{\partial z_y} = \sum_{x \in V} \frac{\partial H}{\partial p_x} \frac{\partial p_x}{\partial z_y}$$
First, the derivative of entropy with respect to probability $p_x$ is:
$$\frac{\partial H}{\partial p_x} = - \frac{\partial}{\partial p_x} (p_x \ln p_x) = -(1 + \ln p_x)$$
Substituting this and the softmax Jacobian into the chain rule summation:
$$\frac{\partial H}{\partial z_y} = \sum_{x \in V} -(1 + \ln p_x) \cdot p_x (\delta_{xy} - p_y)$$
Distributing the terms:
\begin{align*}
    \frac{\partial H}{\partial z_y} &= - \left[ \sum_{x \in V} (1 + \ln p_x) p_x \delta_{xy} - \sum_{x \in V} (1 + \ln p_x) p_x p_y \right] \\
    &= - \left[ p_y(1 + \ln p_y) - p_y \sum_{x \in V} p_x(1 + \ln p_x) \right]
\end{align*}
Expanding the summation $\sum p_x(1 + \ln p_x) = \sum p_x + \sum p_x \ln p_x$.
\begin{align*}
\frac{\partial H}{\partial z_y} &= - \left[ p_y + p_y \ln p_y - p_y (1 - H) \right] \\
&= - \left[ p_y + p_y \ln p_y - p_y + p_y H \right] \\
&= - p_y (\ln p_y + H)
\end{align*}
Expressing this in vector notation yields:
$$\nabla_z H = - \mathbf{p} \odot (\ln \mathbf{p} + H \cdot \mathbf{1})$$
The standard Policy Gradient loss for a selected action $a$ is $L(\theta) \approx \hat{A} \ln p_a$. The gradient with respect to logits is a known standard result involving the one-hot vector $\mathbf{e}_a$:$$\nabla_z L = \hat{A} (\mathbf{e}_a - \mathbf{p})$$
We now compute the dot product between the two gradients derived above. This metric indicates alignment between the learning signal and the direction of entropy growth.
\begin{align*}
\langle \nabla_z L, \nabla_z H \rangle &= \left( \hat{A} (\mathbf{e}_a - \mathbf{p}) \right)^\top \left( - \mathbf{p} \odot (\ln \mathbf{p} + H \cdot \mathbf{1}) \right) \\
&= -\hat{A} (\mathbf{e}_a - \mathbf{p})^\top \left( \mathbf{p} \odot (\ln \mathbf{p} + H \cdot \mathbf{1}) \right)\\
&= -\hat{A} \left[ p_a (\ln p_a + H) - \sum_{x \in V} p_x^2 (\ln p_x + H) \right]
\end{align*}

In our training traces, the sign of the token-specific approximation agrees with the full expression for 91.7\% of sampled updates on average, supporting its empirical use as a diagnostic criterion.

\subsection{Proof of \cref{eq:7}}
\label{appendix:proof2}
\textbf{\cref{eq:7}:} Considering the gradient of the objective for a specific sampled token $a$ with respect to the logit of a generic token $x$ (denoted as $z_x$):
\begin{equation*}
    \begin{split}
        \frac{\partial (\ln \pi_\theta(a \mid s) \cdot \hat{A})}{\partial z_x} &= \frac{\partial \pi_\theta(a \mid s)}{\partial z_x} \cdot \frac{\hat{A}}{\pi_\theta(a \mid s)} \\
        & \hspace{-50pt} =\begin{cases} (1 - \pi_\theta(a \mid s)) \cdot \hat{A} & \text{if } x = a \quad  \\ -\pi_\theta(x \mid s) \cdot \hat{A} & \text{otherwise}  \end{cases} 
    \end{split}
\end{equation*}
\textit{Proof adapted from \cite{gao2025softadaptivepolicyoptimization}.} 
$$\begin{aligned}
\frac{\partial (\ln \pi_\theta(a \mid s) \cdot \hat{A})}{\partial z_x} &= \frac{\partial \pi_\theta(a \mid s)}{\partial z_x} \cdot \frac{\hat{A}}{\pi_\theta(a \mid s)} \\
&= \frac{\mathds{1}(x = a) \exp(z_a) \sum_{x'} \exp(z_{x'}) - \exp(z_a) \exp(z_x)}{\left(\sum_{x'} \exp(z_{x'})\right)^2} \cdot \frac{\hat{A}}{\pi_\theta(a \mid s)} \\
&= \begin{cases} (1 - \pi_\theta(a \mid s)) \cdot \hat{A} & \text{if } x = a \qquad \text{sampled token}\\ -\pi_\theta(x \mid s) \cdot \hat{A} & \text{otherwise} \qquad  \text{unsampled token}\end{cases}
\end{aligned}$$

\section{Experimental Setup}

\label{Appendix:exp-setup}
\subsection{Models and Datasets}
\label{datas}
In the preliminary exploration experiments, we used the \textbf{Qwen2.5-Math-7B} model. In the benchmark evaluation experiments, we utilized \textbf{Qwen2.5-Math-7B} and \textbf{Qwen2.5-7B} as the base policy model. The model is trained using the \texttt{DAPO-Math-17k} dataset. For evaluation and validation, we primarily report performance on the AIME24 \cite{aime24}, AIME25 \cite{aime25}, AMC \cite{lightman2023let}, MATH-500 \cite{lightman2023let}, GSM8k \cite{cobbe2021trainingverifierssolvemath}, and Olympiad \cite{lightman2023let} benchmarks.  To further examine the generality of our method, we additionally evaluate non-mathematical capabilities on GPQA-Diamond \cite{rein2023gpqa}, LiveCodeBench \cite{jain2024livecodebench}, and MMLU-Redux \cite{gema2024mmlu} using Qwen2.5-Math-7B, and conduct experiments on Phi-4-14B \cite{abdin2024phi}, a non-Qwen model family, trained for 200 steps. To support complex reasoning tasks, we configure the maximum sequence lengths to 4096 tokens for Qwen2.5-Math-7B and 8192 tokens for Qwen2.5-7B and Phi-4-14B.

\textbf{Assets, Licenses, and Terms of Use.}
All models and datasets used in this work are publicly available research assets. 
The Qwen2.5-Math-7B and Qwen2.5-7B models are released by the Qwen team on Hugging Face; Qwen2.5-7B is distributed under the Apache-2.0 license, and the Qwen2.5-Math model card identifies Qwen2.5-7B as its base model and provides the corresponding public model release and citation information. 
Phi-4-14B is released by Microsoft under the MIT license according to its official model card. 
The training dataset DAPO-Math-17k is publicly released on Hugging Face under the Apache-2.0 license.

For evaluation, we use only publicly available benchmark datasets and follow their intended research/evaluation use. 
GSM8K is released under the MIT license. 
MATH-500 is a subset of the MATH benchmark; the original MATH dataset is released under the MIT license. 
GPQA is released with an MIT-licensed repository, and MMLU-Redux is released under the CC-BY-4.0 license. 
OlympiadBench is released under a Creative Commons Attribution-NonCommercial license according to its official repository. 
For AIME and AMC, which are competition-style benchmarks commonly used for academic evaluation, we use them strictly for non-commercial research evaluation and provide citations to the original benchmark sources where applicable. 
LiveCodeBench is a continuously updated benchmark constructed from public programming contest problems; we use it only for evaluation and do not redistribute the original contest data beyond what is permitted by the benchmark release.

\subsection{Training Configuration}
We employ GRPO as the advantage estimator. The model is trained for a total of 400 steps with a global batch size of 512. During the rollout phase, we sample $N=8$ responses per prompt to estimate the baseline and advantages. The optimization uses the AdamW optimizer with the following hyperparameters:
\begin{itemize}
    \item \textbf{Learning Rate:} $1 \times 10^{-6}$.
    \item \textbf{Weight Decay:} 0.1.
    \item \textbf{Gradient Clipping:} 1.0.
    \item \textbf{KL penalty coefficient:} The KL penalty coefficient ($\beta_{KL}$) is set to 0.0.
\end{itemize}
For the dynamic linear parameters of the upper clipping threshold and lower clipping threshold, we calibrated the upper clipping threshold to exceed $\epsilon_{high}$ in the low-probability regime while remaining lower in the high-probability regime (the same applies to the lower clipping threshold). The linear slope and intercept of the dynamic upper clipping threshold are set to $-0.25$ and $0.5$, and the linear slope and intercept of the dynamic lower clipping threshold are set to $-0.13$ and $0.3$. Experiments have shown that this setting has broad generalizability.

\subsection{Implementation Details}
Experiments are conducted on a single node equipped with 8 $\times$ H100 GPUs. To maximize training throughput and memory efficiency, we implement a hybrid parallelism strategy using the \texttt{verl} \cite{sheng2024hybridflow} framework:
\begin{itemize}
    \item \textbf{Inference:} We utilize the \texttt{vLLM} \cite{kwon2023efficient} engine with a tensor model parallelism size of 4. We employ a sampling temperature of 1.0 and Top-$p=1.0$.
    \item \textbf{Training:} The actor model is trained using Fully Sharded Data Parallel with parameter and optimizer offloading enabled. We additionally utilize Ulysses sequence parallelism with a size of 4.
\end{itemize}
\subsection{Computational Cost Analysis}
We measure the training time for some methods over 400 training steps as shown in Table \ref{tab:training_time}.
\begin{table}[h]
    \centering
    \caption{Training time comparison for 400 steps on 8$\times$ 80GB H100 GPUs. Times are formatted as Hours:Minutes or Days:Hours:Minutes.}
    \label{tab:training_time}
    \begin{tabular}{lcc}
        \toprule
        \textbf{Method} & \textbf{Qwen2.5-7B-Math} & \textbf{Qwen2.5-7B} \\
        \midrule
        GRPO (Baseline) & 17h 24m & 1d 03h 18m \\
        Clip-Higher & 18h 01m & 1d 05h 47m \\
        Clip-Cov & 18h 41m & 1d 02h 15m \\
        \midrule
        \textbf{Ours-ID} & 18h 36m & 1d 03h 55m \\
        \textbf{Ours-DID} & 17h 44m & 1d 02h 30m \\
        \textbf{Ours-OD} & 18h 14m & 1d 03h 49m \\
        \bottomrule
    \end{tabular}
\end{table}

\section{Evaluation Setup}
\label{Appendix:eval-setup}
\subsection{Implementation Details}
Our evaluation pipeline utilizes the \texttt{EvalScope} \cite{evalscope_2024} framework. Inference is served using \texttt{lmdeploy} \cite{zhang2025efficient} with a PyTorch backend. The serving infrastructure is distributed across 8 GPUs. The maximum output length of the model is consistent with that during training.

We utilize a unified sampling configuration across all benchmarks to ensure consistency. The sampling parameters are detailed in Table \ref{tab:hyperparams}.

\begin{table}[h]
\centering
\caption{Inference Sampling Hyperparameters}
\label{tab:hyperparams}
\begin{tabular}{lc}
\toprule
\textbf{Parameter} & \textbf{Value} \\
\midrule
Temperature & 0.7 \\
Top-$p$ & 0.8 \\
Top-$k$ & 20 \\
Batch Size & 256 \\
\bottomrule
\end{tabular}
\end{table}

\subsection{Benchmarks and Metrics}
We evaluate the models on a suite of mathematical reasoning benchmarks. The evaluation metric is \texttt{mean\_and\_pass\_at\_k}. The number of samples generated per problem ($N$) varies by dataset scale:

\begin{itemize}
    \item \textbf{32 samples:} AMC, AIME 2024, AIME 2025.
    \item \textbf{16 samples:} GPQA-Diamond.
    \item \textbf{4 samples:} MATH-500, OlympiadBench (Subset: \texttt{OE\_TO\_maths\_en\_COMP}).
    \item \textbf{2 samples:} GSM8K.
    \item \textbf{1 sample:} LiveCodeBench, MMLU-Redux.
\end{itemize}

\subsection{Prompt Template}
For all mathematical reasoning tasks, we employ a standard chain-of-thought prompt designed to enforce a specific output format. The template used is:

\begin{rqbox}[title={Prompt}]
\vspace{-10pt}
\{question\} 

Please reason step by step, and put your final answer within \textbackslash boxed\{\{\}\}.
\vspace{-5pt}
\end{rqbox}

\section{Other Experimental Results}
\label{appendix:experimental}
\subsection{Analysis of the Curves of Entropy and Avg Clipping Threshold}

\begin{figure}[h] 
\centering
\includegraphics[width=1\linewidth]{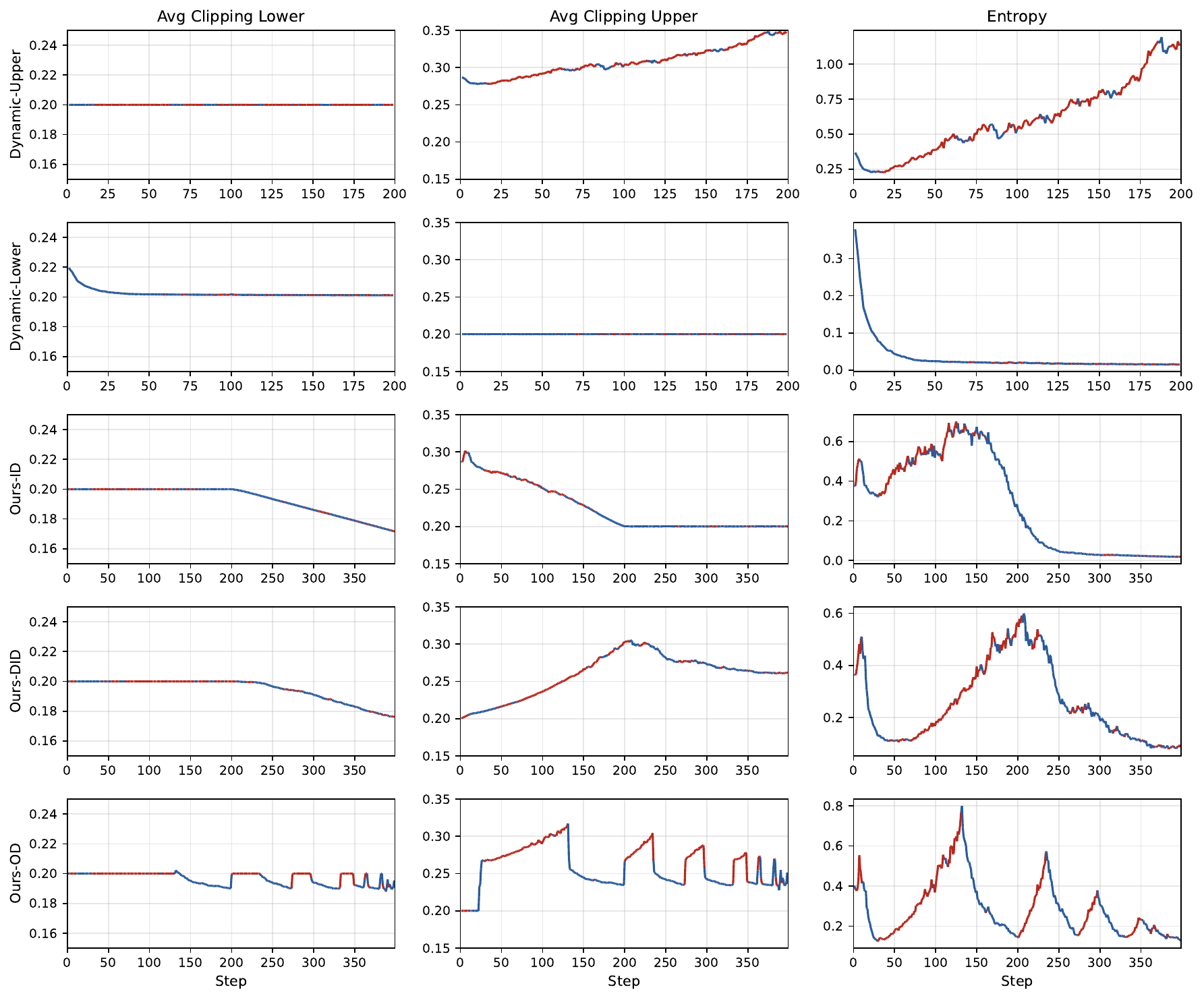} 
\caption{Comparison of entropy change and avg clipping threshold change for the Qwen2.5-Math-7B model}
\label{fig:avgclipping1}
\vspace{-10pt}
\end{figure}

\begin{figure}[h] 
\centering
\includegraphics[width=1\linewidth]{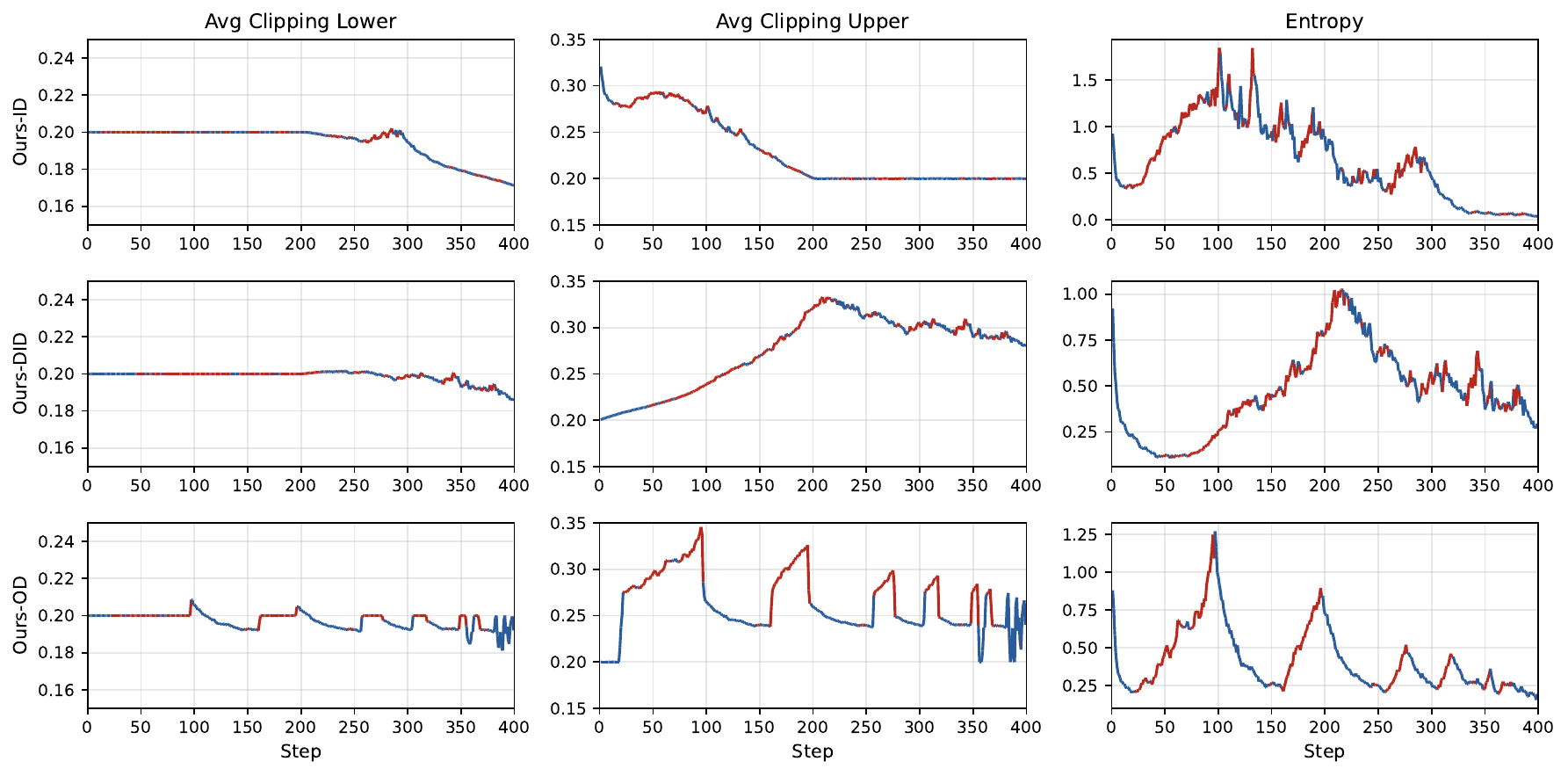} 
\caption{Comparison of entropy change and avg clipping threshold change for the Qwen2.5-7B model}
\label{fig:avgclipping2}
\vspace{-10pt}
\end{figure}

We saved the entropy and average clipping threshold curves during the training process of the Dynamic Clipping Lower threshold, Dynamic Clipping Upper threshold, Ours-ID, Ours-DID, and Ours-OD methods for Qwen2.5-Math-7B. As shown in \cref{fig:avgclipping1}, the entropy and average clipping threshold curves during the training process of the Ours-ID, Ours-DID, and Ours-OD methods for Qwen2.5-7B are presented in \cref{fig:avgclipping2}.

It can be seen that when the model's average clipping upper threshold is larger, the entropy of the model tends to show an increasing trend, and when the clipping upper threshold decreases and the clipping lower threshold also decreases, the model's entropy shows an decreasing trend. This also verifies the effectiveness of our adjustment mechanism.

\subsection{Analysis of Clipping Function}
\label{clipping-function}

\paragraph{Justification of the linear clipping schedule.}
We adopt a linear probability-dependent clipping schedule for both theoretical and practical considerations.
First, our theoretical analysis suggests that the clipping radius $\epsilon$ should be negatively correlated with the reference probability $\pi_{\theta}(y_t \mid x, y_{<t})$.
A linear schedule is the simplest functional family that satisfies this monotonicity requirement, while introducing minimal additional complexity into large-scale LLM training.
Compared with more complex nonlinear alternatives, it is also easier to calibrate and less prone to unstable threshold variations.

Second, the linear form provides favorable numerical behavior.
Nonlinear decay functions may decrease the clipping threshold too aggressively for medium- and high-probability tokens, thereby excessively shrinking the effective trust region.
In contrast, a linear schedule induces a constant and predictable decay rate, leading to a smoother contraction of the trust region across probability levels.

To further examine whether the linear form is a reasonable design choice, we conduct an ablation study comparing it with an exponential alternative,
\[
\epsilon(p) = \alpha \exp(-\lambda p),
\]
where the parameters are calibrated such that the exponential schedule matches the same boundary values as the linear schedule.
The comparison is performed under the upper-bound-only dynamic clipping setting.
As shown in Table~\ref{tab:linear_vs_nonlinear_passk}, the linear schedule consistently outperforms the nonlinear variant across all Pass@$K$ metrics.

\begin{table}[h]
\centering
\caption{Comparison between linear and nonlinear clipping schedules under the upper-bound-only dynamic clipping setting.}
\label{tab:linear_vs_nonlinear_passk}
\begin{tabular}{lcccc}
\toprule
\textbf{Form} & \textbf{Pass@1} & \textbf{Pass@8} & \textbf{Pass@16} & \textbf{Pass@32} \\
\midrule
Linear     & \textbf{63.11} & \textbf{79.21} & \textbf{82.04} & \textbf{84.78} \\
Nonlinear  & 62.50          & 77.26          & 79.85          & 82.61          \\
\bottomrule
\end{tabular}
\end{table}

We also evaluate the two schedules in our methods Ours-ID and Ours-OD on Qwen2.5-Math-7B.
As reported in Table~\ref{tab:linear_vs_nonlinear_id_od}, the linear schedule achieves stronger or comparable performance on most benchmarks.

\begin{table}[h]
\centering
\caption{ID/OD comparison between linear and nonlinear clipping schedules on Qwen2.5-Math-7B.}
\label{tab:linear_vs_nonlinear_id_od}
\begin{tabular}{lcccccc}
\toprule
\textbf{Method} & \textbf{AIME24} & \textbf{AIME25} & \textbf{AMC} & \textbf{MATH-500} & \textbf{GSM8K} & \textbf{Olympiad} \\
\midrule
Linear-ID     & \textbf{33.1} & \textbf{18.1} & \textbf{58.9} & \textbf{84.5}  & \textbf{93.7}  & \textbf{48.3} \\
Nonlinear-ID  & 30.94         & 15.62         & 57.30         & 81.30          & 91.96          & 46.36          \\
Linear-OD     & \textbf{31.1} & \textbf{18.3} & \textbf{58.3} & 83.0           & 92.4           & \textbf{47.5} \\
Nonlinear-OD  & 30.5          & 17.50         & 56.16         & \textbf{83.95} & \textbf{92.42} & 46.66          \\
\bottomrule
\end{tabular}
\end{table}

These results indicate that the effectiveness of probability-dependent clipping primarily comes from adapting the trust-region constraint according to token probability, rather than from using a highly expressive functional form.
The simple linear schedule therefore provides a strong trade-off between theoretical consistency, numerical stability, and empirical performance.

Nevertheless, we acknowledge that a fixed functional form with fixed hyperparameters may not fully capture the flexibility intended by probability-dependent clipping.
Exploring more diversified, adaptive, or learnable clipping schedules is an important direction for future work.

\subsection{Analysis of Clipping Probability Curve}
\begin{figure}[H] 
\centering
\includegraphics[width=1\linewidth]{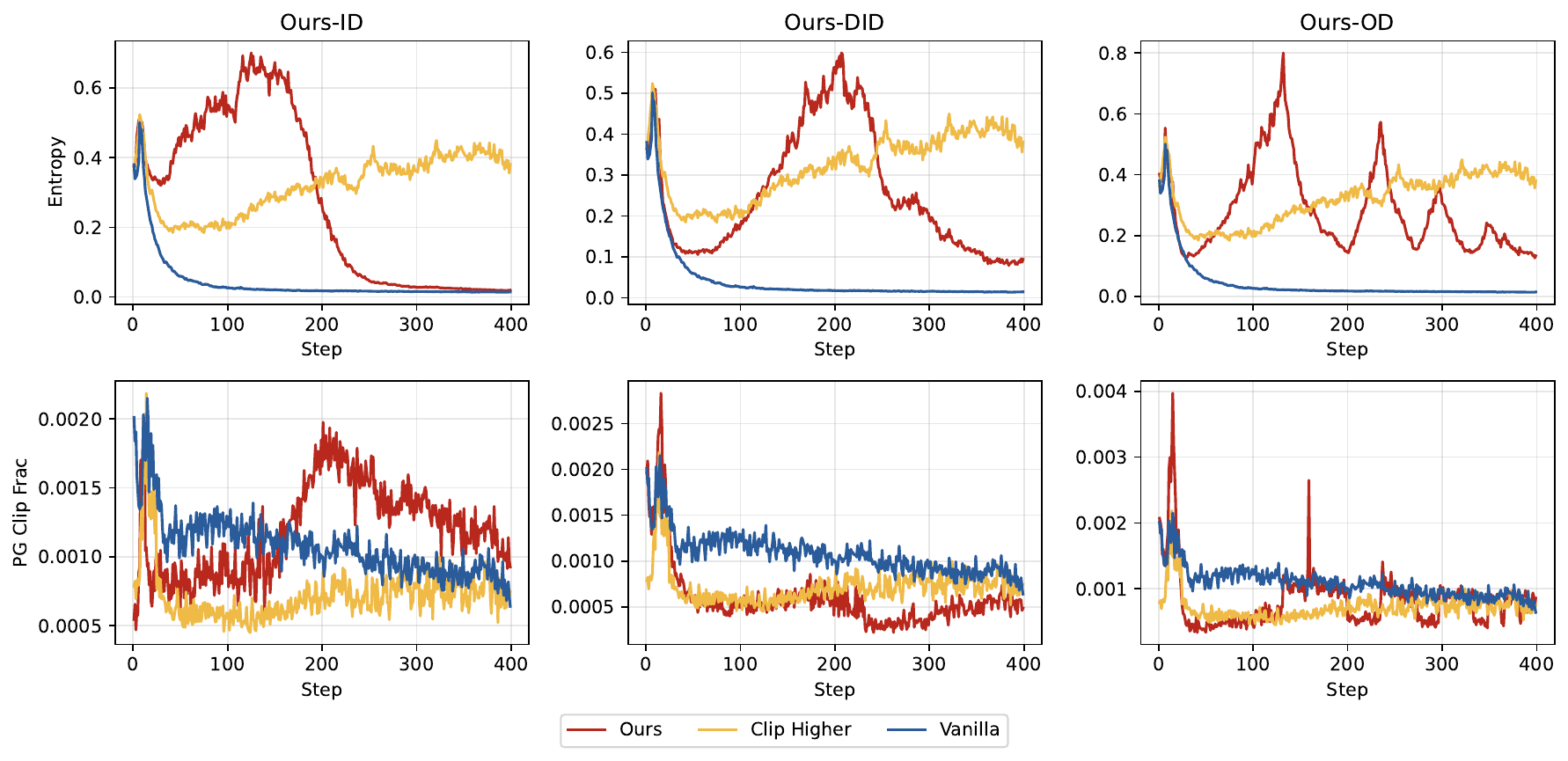} 
\caption{Graph of Model Entropy and Average Token Clipping Probability}
\label{fig:clipfrac}
\vspace{-10pt}
\end{figure}

The clipping probability of the model refers to the proportion of tokens that are clipped during the model's training process. This proportion reflects the number of tokens affected by the clipping mechanism during the model's training.

As can be seen from \cref{fig:clipfrac}, our entropy control strategy is effectively reflected in the clipping ratio during the model's training process. For example, in the early stages of training, the token clipping probability of the Ours-ID method is low, and at this time, the entropy is in a period of increase. In the later stages of training, the token clipping probability of the Ours-ID method is high, and at this time, it is in a period of entropy decrease. In the Ours-OD method, this characteristic is reflected in the fact that the entropy also fluctuates throughout the training process.

\subsection{Experiment on Replacing Dynamic Clipping Threshold with Clip-Higher and Clip-Lower}
\begin{figure}[h] 
\centering
\includegraphics[width=1\linewidth]{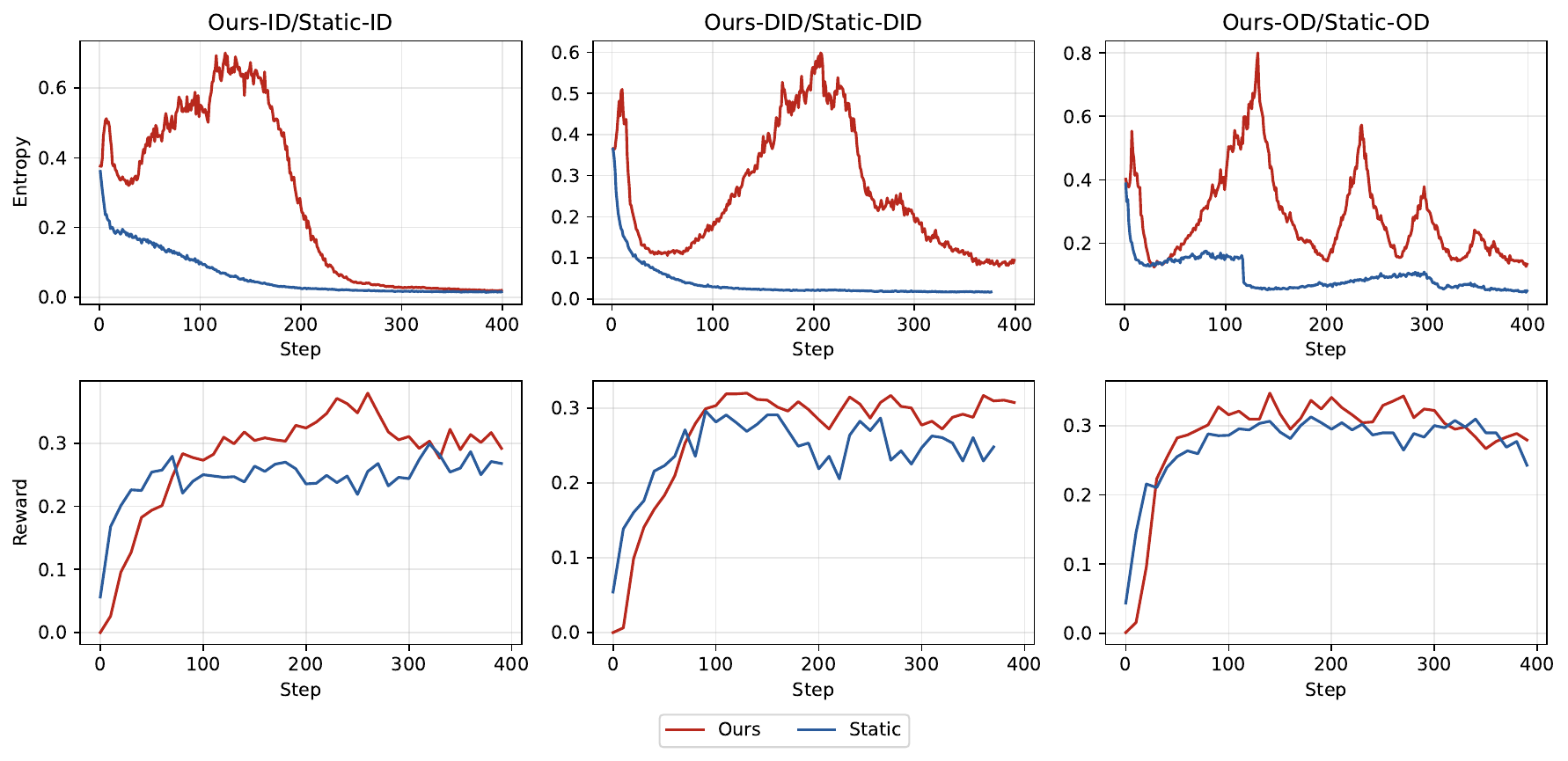} 
\caption{Experiment on Replacing Dynamic Clipping Threshold with Clip-Higher and Clip-Lower}
\label{fig:dapo}
\vspace{-10pt}
\end{figure}

If we do not use dynamic upper clipping threshold and dynamic lower clipping threshold, but instead use Clip-Higher and Clip-Lower as the adjustment mechanism for entropy increase and decrease, and apply them to the entropy control strategy in our \cref{methdology3}, the final results are shown in \cref{fig:dapo}.

We can see that although Clip-Higher and Clip-Lower are effective means of controlling entropy increase and decrease, they may affect the performance of the model because they adopt a one-size-fits-all approach to tokens in different regions without dynamic adjustment. On the other hand, the entropy increase and decrease of the model cannot be effectively and precisely controlled by Clip-Higher and Clip-Lower, and we fail to observe the intended increase-then-decrease or decrease-increase-decrease entropy patterns. It still faces the problem of entropy collapse.

\section{Implementation Code Example}
\label{code}
\subsection{Ours-ID/DID}
Compared with GRPO, Ours-ID/DID preserves the original clipped policy objective but replaces the fixed scalar clipping range with annealed token-dependent clipping bounds. Specifically, the upper and lower clipping ranges are parameterized as functions of the detached current policy probability, where the coefficients are linearly annealed during training. We apply stop-gradient to this probability so that gradients do not flow through the dynamic clipping bounds.

\begin{lstlisting}[
language=Python,
caption={Core implementation of Ours-ID based on minimal modification to GRPO.},
escapeinside={(*@}{@*)}
]
def compute_policy_loss_ours_id(
    old_log_prob, log_prob, advantages, response_mask, config
):
    # ===== GRPO: likelihood ratio =====
    log_ratio = torch.clamp(log_prob - old_log_prob, -20.0, 20.0)
    ratio = torch.exp(log_ratio)

    # ===== Ours-ID/DID: dynamic token-level clipping =====
    progress = torch.clamp(
        config.current_step / config.total_steps, 0.0, 1.0
    )

    a = lerp(config.a_start, config.a_end, progress)
    b = lerp(config.b_start, config.b_end, progress)
    c = lerp(config.c_start, config.c_end, progress)
    d = lerp(config.d_start, config.d_end, progress)

    prob = torch.exp(log_prob).detach()

    (*@\textcolor{red}{\# GRPO uses fixed scalar clipping bounds}@*)
    (*@\textcolor{red}{\# clip\_low  = config.clip\_ratio}@*)
    (*@\textcolor{red}{\# clip\_high = config.clip\_ratio}@*)

    (*@\textcolor{blue}{\# Ours-ID uses annealed probability-dependent bounds}@*)
    (*@\textcolor{blue}{\# prob is detached; thresholds are not optimized directly}@*)
    (*@\textcolor{blue}{clip\_high = a * prob + b}@*)
    (*@\textcolor{blue}{clip\_low  = c * prob + d}@*)

    # ===== clipped policy objective =====
    loss_unclipped = -advantages * ratio

    loss_clipped = -advantages * torch.clamp(
        ratio,
        1.0 - clip_low,
        1.0 + clip_high
    )

    loss = torch.maximum(loss_unclipped, loss_clipped)

    return masked_mean(loss, response_mask)
\end{lstlisting}

\subsection{Ours-OD}
Ours-OD preserves the GRPO clipped objective, but replaces the fixed clipping range with an entropy-driven state machine. When entropy is too high, it restricts probability increases; when entropy is too low, it relaxes the upper clipping bound to encourage exploration.

\begin{lstlisting}[
language=Python,
caption={Core implementation of Ours-OD based on minimal modification to GRPO.},
escapeinside={(*@}{@*)}
]
def compute_policy_loss_ours_od(
    old_log_prob, log_prob, advantages, response_mask, entropy, config
):
    # ===== GRPO: likelihood ratio =====
    log_ratio = torch.clamp(log_prob - old_log_prob, -20.0, 20.0)
    ratio = torch.exp(log_ratio)

    # ===== Ours-OD: entropy-driven clipping state =====
    mean_entropy = masked_mean(entropy, response_mask)

    if not hasattr(config, "initial_entropy"):
        config.initial_entropy = mean_entropy
        config.clip_state = "normal"

    progress = torch.clamp(
        config.current_step / config.total_steps, 0.0, 1.0
    )

    H_upper = config.initial_entropy * lerp(
        config.h_upper_start, config.h_upper_end, progress
    )
    H_lower = config.initial_entropy * config.h_lower

    if mean_entropy > H_upper:
        config.clip_state = "high"
    elif mean_entropy < H_lower:
        config.clip_state = "low"

    prob = torch.exp(log_prob).detach()

    (*@\textcolor{red}{\# GRPO uses fixed symmetric clipping bounds}@*)
    (*@\textcolor{red}{\# clip\_high = config.clip\_ratio}@*)
    (*@\textcolor{red}{\# clip\_low  = config.clip\_ratio}@*)

    (*@\textcolor{blue}{\# Ours-OD adapts clipping bounds according to entropy state}@*)
    (*@\textcolor{blue}{\# prob is detached; thresholds are not optimized directly}@*)
    if config.clip_state == "normal":
        clip_high = config.clip_ratio
        clip_low  = config.clip_ratio

    elif config.clip_state == "high":
        (*@\textcolor{blue}{\# high entropy: restrict probability increases}@*)
        clip_high = config.clip_ratio
        clip_low  = c * prob + d

    else:  # low entropy
        (*@\textcolor{blue}{\# low entropy: encourage probability increases}@*)
        clip_high = a * prob + b
        clip_low  = config.clip_ratio

    # ===== clipped policy objective =====
    loss_unclipped = -advantages * ratio

    loss_clipped = -advantages * torch.clamp(
        ratio,
        1.0 - clip_low,
        1.0 + clip_high
    )

    loss = torch.maximum(loss_unclipped, loss_clipped)

    return masked_mean(loss, response_mask)
\end{lstlisting}

\section{Broader Impacts}
\label{impacts}
This work contributes to the development of more stable and effective RL methods for LLMs, particularly in reasoning-oriented settings with verifiable rewards. By mitigating entropy collapse and improving the controllability of policy entropy during training, the proposed method may help build models that maintain better exploration, avoid premature overconfidence, and achieve stronger reasoning performance. These improvements can benefit applications such as mathematical problem solving, coding assistance, scientific reasoning, and other domains where reliable multi-step reasoning is valuable.

At the same time, improving the training stability and reasoning ability of LLMs may also amplify existing risks associated with advanced language models. Stronger reasoning models could be misused to automate harmful tasks, generate more convincing misleading content, or assist in activities requiring planning and technical expertise. In addition, entropy-control strategies that improve model confidence and convergence may increase the risk of overconfident incorrect outputs if models are deployed without appropriate evaluation and safeguards. Therefore, practical deployment of models trained with such methods should include careful benchmark evaluation, safety testing, monitoring for misuse, and domain-specific risk assessment. This paper focuses on an optimization mechanism and does not release a new high-risk model or dataset, but the broader implications should be considered when applying the method to large-scale model training or public-facing systems.

\newpage


\newpage
\section*{NeurIPS Paper Checklist}

\begin{enumerate}

\item {\bf Claims}
    \item[] Question: Do the main claims made in the abstract and introduction accurately reflect the paper's contributions and scope?
    \item[] Answer: \answerYes{} 
    \item[] Justification: The abstract and Introduction state the paper's scope and contributions: analyzing entropy dynamics from a Gradient-Preserving Clipping perspective, proposing dynamic clipping thresholds, and evaluating ID, DID, and OD entropy-control strategies. These claims are supported by the theoretical analysis in Section~\ref{methdology1}, the method in Section~\ref{methdology2}--\ref{methdology3}, and the experiments in Section~\ref{analysis1}.

\item {\bf Limitations}
    \item[] Question: Does the paper discuss the limitations of the work performed by the authors?
    \item[] Answer: \answerYes{} 
    \item[] Justification: We discusses limitations of the fixed linear clipping schedule and notes that more diversified, adaptive, or learnable clipping schedules remain future work in Section~\ref{clipping-function}. The experiments also analyze sensitivity to phase ratios and clipping-function choices in Section~\ref{analysis3}.

\item {\bf Theory assumptions and proofs}
    \item[] Question: For each theoretical result, does the paper provide the full set of assumptions and a complete (and correct) proof?
    \item[] Answer: \answerYes{} 
    \item[] Justification: The assumptions used in the entropy-gradient analysis are stated in Section~\ref{methdology1}, including the token-level surrogate objective, softmax-logit parameterization, and advantage-sign cases. Detailed derivations and proofs for the key equations are provided in Section~\ref{appendix:proof1} and Section~\ref{appendix:proof2}.

    \item {\bf Experimental result reproducibility}
    \item[] Question: Does the paper fully disclose all the information needed to reproduce the main experimental results of the paper to the extent that it affects the main claims and/or conclusions of the paper (regardless of whether the code and data are provided or not)?
    \item[] Answer: \answerYes{} 
    \item[] Justification: We disclose the models, datasets, benchmarks, baselines, training steps, batch size, rollout count, optimizer settings, clipping schedules, sequence lengths, evaluation framework, sampling settings, and compute setup in Section~\ref{Appendix:exp-setup} and Section~\ref{Appendix:eval-setup}. The method formulas in Section~\ref{methdology2}--\ref{methdology3} specify the algorithmic changes needed to reproduce the main results.

\item {\bf Open access to data and code}
    \item[] Question: Does the paper provide open access to the data and code, with sufficient instructions to faithfully reproduce the main experimental results, as described in supplemental material?
    \item[] Answer: \answerYes{} 
    \item[] Justification: We describe the experimental setup in detail. The training dataset, models, frameworks, and benchmarks are identified in Section~\ref{Appendix:exp-setup} and Section~\ref{Appendix:eval-setup}. We also showcase code examples of our core control strategies (including Ours-ID and Ours-OD) in Section \ref{code}.

\item {\bf Experimental setting/details}
    \item[] Question: Does the paper specify all the training and test details (e.g., data splits, hyperparameters, how they were chosen, type of optimizer) necessary to understand the results?
    \item[] Answer: \answerYes{} 
    \item[] Justification: The experimental setup in the main text specifies the models, training data, benchmarks, baselines, learning rate, rollout count, global batch size, and response lengths. Additional training and evaluation details, including optimizer hyperparameters, clipping-parameter settings, inference framework, sampling parameters, and prompt template, are provided in Section~\ref{Appendix:exp-setup} and Section~\ref{Appendix:eval-setup}.

\item {\bf Experiment statistical significance}
    \item[] Question: Does the paper report error bars suitably and correctly defined or other appropriate information about the statistical significance of the experiments?
    \item[] Answer: \answerYes{} 
    \item[] Justification: We report averaged results using the \texttt{mean\_and\_pass\_at\_k} metric, with multiple sampled responses per problem for stochastic evaluation and found error bar is relatively
small. Our experimental results have high credibility. The number of samples used for each benchmark is specified in Section~\ref{Appendix:eval-setup}.

\item {\bf Experiments compute resources}
    \item[] Question: For each experiment, does the paper provide sufficient information on the computer resources (type of compute workers, memory, time of execution) needed to reproduce the experiments?
    \item[] Answer: \answerYes{} 
    \item[] Justification: The appendix reports that experiments use a single node with 8$\times$ 80GB H100 GPUs and includes training-time comparisons for 400-step runs in Table~\ref{tab:training_time}.
    
\item {\bf Code of ethics}
    \item[] Question: Does the research conducted in the paper conform, in every respect, with the NeurIPS Code of Ethics \url{https://neurips.cc/public/EthicsGuidelines}?
    \item[] Answer: \answerYes{} 
    \item[] Justification: The research uses existing public models and benchmarks for algorithmic study, does not involve human subjects, and preserves submission anonymity. We have reviewed the NeurIPS Code of Ethics and are not aware of deviations from it.

\item {\bf Broader impacts}
    \item[] Question: Does the paper discuss both potential positive societal impacts and negative societal impacts of the work performed?
    \item[] Answer: \answerYes{} 
    \item[] Justification: The paper motivates positive impacts through improved RLVR training stability and reasoning performance, with the specific impacts elaborated in Section \ref{impacts}.
    
\item {\bf Safeguards}
    \item[] Question: Does the paper describe safeguards that have been put in place for responsible release of data or models that have a high risk for misuse (e.g., pre-trained language models, image generators, or scraped datasets)?
    \item[] Answer: \answerNA{} 
    \item[] Justification: We propose and evaluate an optimization method and does not announce the release of a new pretrained language model, image generator, or scraped dataset with high misuse risk. The experiments are conducted on existing models and public benchmarks.

\item {\bf Licenses for existing assets}
    \item[] Question: Are the creators or original owners of assets (e.g., code, data, models), used in the paper, properly credited and are the license and terms of use explicitly mentioned and properly respected?
    \item[] Answer: \answerYes{} 
    \item[] Justification: We cite the existing models, datasets, benchmarks, and software frameworks used in the experiments, including Qwen, Phi, DAPO-Math, EvalScope, lmdeploy, vLLM, and verl. We explicitly state the licenses or terms of use for these assets in Section \ref{datas}.

\item {\bf New assets}
    \item[] Question: Are new assets introduced in the paper well documented and is the documentation provided alongside the assets?
    \item[] Answer: \answerNA{} 
    \item[] Justification: We do not introduce or release a new dataset, benchmark, model checkpoint, or software asset as a primary contribution. The contribution is an entropy-control mechanism and training strategy evaluated using existing assets.

\item {\bf Crowdsourcing and research with human subjects}
    \item[] Question: For crowdsourcing experiments and research with human subjects, does the paper include the full text of instructions given to participants and screenshots, if applicable, as well as details about compensation (if any)? 
    \item[] Answer: \answerNA{} 
    \item[] Justification: The work does not involve crowdsourcing, user studies, or research with human subjects. All experiments use existing datasets, models, and automated benchmark evaluations.

\item {\bf Institutional review board (IRB) approvals or equivalent for research with human subjects}
    \item[] Question: Does the paper describe potential risks incurred by study participants, whether such risks were disclosed to the subjects, and whether Institutional Review Board (IRB) approvals (or an equivalent approval/review based on the requirements of your country or institution) were obtained?
    \item[] Answer: \answerNA{} 
    \item[] Justification: The work does not involve crowdsourcing or human-subject research, so IRB approval or an equivalent review is not applicable. The experiments are based on existing machine learning benchmarks.

\item {\bf Declaration of LLM usage}
    \item[] Question: Does the paper describe the usage of LLMs if it is an important, original, or non-standard component of the core methods in this research? Note that if the LLM is used only for writing, editing, or formatting purposes and does \emph{not} impact the core methodology, scientific rigor, or originality of the research, declaration is not required.
    \item[] Answer: \answerYes{} 
    \item[] Justification: The research is about reinforcement learning for LLMs, and the use of LLM base policies is central to the method and experiments. The paper describes the specific LLMs used, including Qwen2.5-Math-7B, Qwen2.5-7B, and Phi-4-14B, in the experimental setup and appendix.

\end{enumerate}

\end{document}